%% file: arxiv_bytedance.tex
\documentclass[]{bytedance}
\usepackage[toc,page,header]{appendix}

\usepackage{minitoc}
\usepackage{amsfonts}
\usepackage{amssymb}
\usepackage{tabularx}
\usepackage{listings}
\usepackage{xcolor}
\usepackage{cancel}

\usepackage{tabulary,multirow,xspace}
\usepackage{makecell}
\usepackage{fixmath,mathtools,nicefrac,mmstyle}
\usepackage{subcaption}
\usepackage{caption}
\usepackage{float}
\usepackage{wrapfig} 
\usepackage[misc]{ifsym} 
\usepackage{colortbl}
\usepackage{lineno}

\usepackage{wrapfig}
\usepackage{multicol}
\usepackage[most]{tcolorbox}
\usepackage{pifont}

\usepackage{color}
\usepackage{colortbl}
\usepackage[dvipsnames, svgnames, x11names]{xcolor}
\usepackage[misc]{ifsym}
\usepackage{soul}
\usepackage{bm}
\usepackage{placeins}
\usepackage{tikz}

\definecolor{codegreen}{rgb}{0,0.6,0}
\definecolor{codegray}{rgb}{0.5,0.5,0.5}
\definecolor{codepurple}{rgb}{0.58,0,0.82}
\definecolor{backcolour}{rgb}{0.95,0.95,0.92}
\definecolor{boxblue}{RGB}{57,89,163}
\definecolor{boxbluebg}{RGB}{230,237,250} 

\lstdefinestyle{mystyle}{
    backgroundcolor=\color{backcolour},   
    commentstyle=\color{codegreen},
    keywordstyle=\color{magenta},
    numberstyle=\tiny\color{codegray},
    stringstyle=\color{codepurple},
    basicstyle=\ttfamily\footnotesize,
    breakatwhitespace=false,         
    breaklines=true,                 
    captionpos=b,                    
    keepspaces=true,                 
    numbers=none,                    
    numbersep=5pt,                  
    showspaces=false,                
    showstringspaces=false,
    showtabs=false,                  
    tabsize=2
}
\definecolor{mygray1}{gray}{.95}
\definecolor{mygray2}{gray}{.9}
\definecolor{mygray3}{gray}{.95}
\usepackage{pifont}

\definecolor{lightgray}{rgb}{0.8, 0.8, 0.8}
\definecolor{lgray}{rgb}{0.66, 0.66, 0.66}

\definecolor{lblu_tab}{RGB}{225, 235, 246}
\definecolor{orange_vitad}{RGB}{222, 131, 68}
\definecolor{blue_vitad}{RGB}{106, 153, 208}
\definecolor{trajectory_green}{RGB}{126, 171, 85}
\definecolor{trajectory_yellow}{RGB}{245, 194, 66}
\definecolor{tab_others}{RGB}{235, 235, 235}
\definecolor{tab_ours}{RGB}{225, 235, 246}
\definecolor{lightblue}{rgb}{0.93, 0.96, 1.0}

\definecolor{whit_tab}{RGB}{255, 255, 255}
\definecolor{gray_tab}{RGB}{246, 246, 246}
\definecolor{oran_tab}{RGB}{252, 242, 237}
\definecolor{blue_tab}{RGB}{227, 240, 251}

\newlength\savewidth
\newcolumntype{x}[1]{>{\centering\arraybackslash}p{#1pt}}

\newcommand{\app}{\raise.17ex\hbox{$\scriptstyle\sim$}}

\usepackage{xcolor}
\usepackage{listings}

\definecolor{commentgreen}{rgb}{0.1, 0.4, 0.1}
\definecolor{keywordblue}{rgb}{0.1, 0.1, 0.7}
\definecolor{stringred}{rgb}{0.7, 0.1, 0.1}

\lstdefinestyle{mystyle}{
    commentstyle=\color{commentgreen},
    keywordstyle=\color{keywordblue},   
    stringstyle=\color{stringred},
    basicstyle=\ttfamily\scriptsize, 
    breaklines=true,
    keepspaces=true,
    showstringspaces=false,
    frame=none,                     
    language=Python, 
}

\newcounter{prompt}
\renewcommand{\theprompt}{\arabic{prompt}}
\newcommand{\prompt}[3]{
\refstepcounter{prompt}
\begin{tcolorbox}[
    colback=lightblue!35, 
    colframe=white!45!black, 
    title={Prompt.~\theprompt:~#1},
    breakable,
]
#2
\label{#3}
\end{tcolorbox}
}

\newcommand{\circled}[1]{\tikz[baseline=(char.base)]{
    \node[shape=circle,draw,inner sep=0pt,minimum size=1em] (char) {#1};}}

\title{\includegraphics[width=0.9cm]{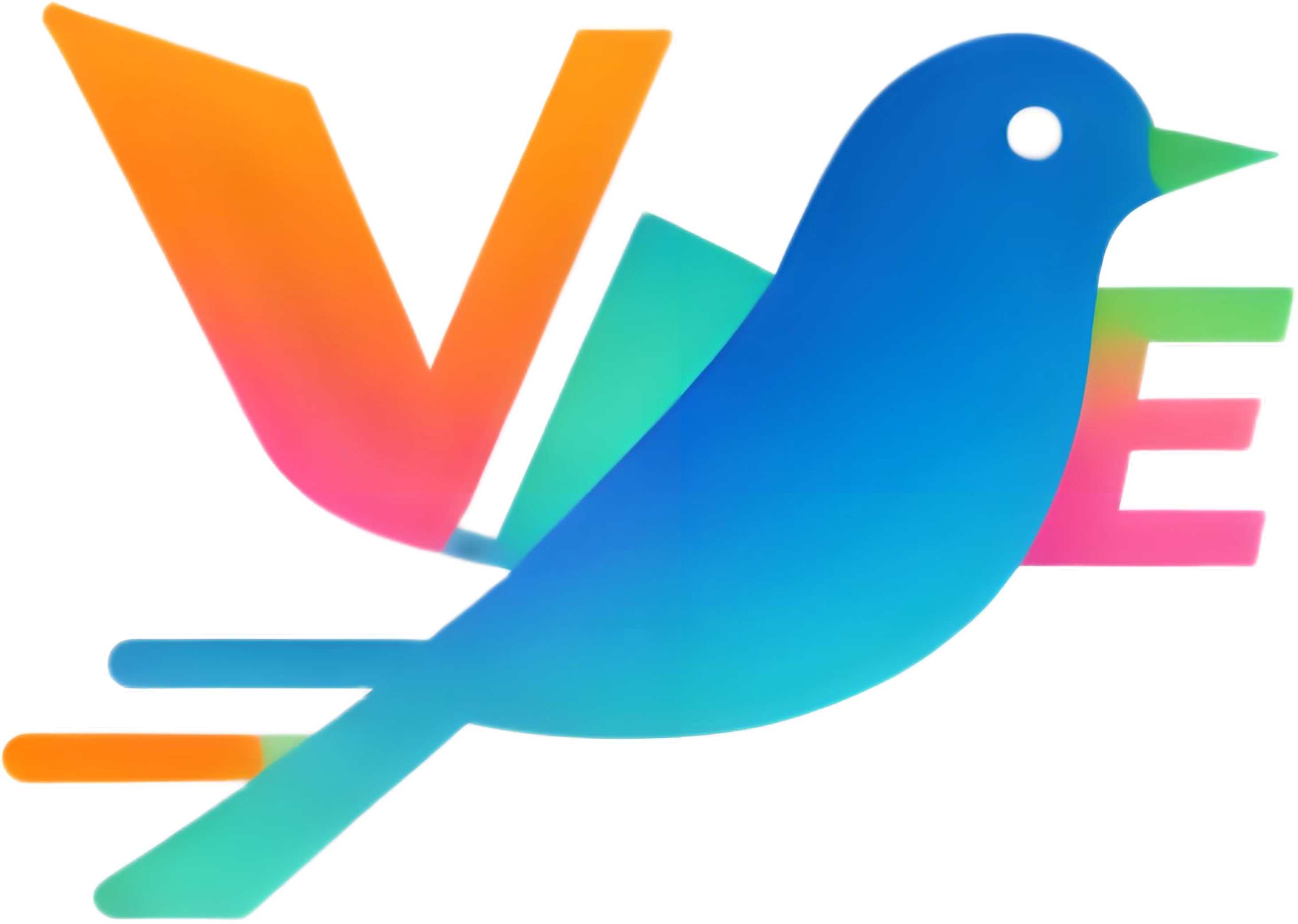} OpenVE-3M: \\A Large-Scale High-Quality Dataset for Instruction-Guided Video Editing}

\author{
\centerline{
Haoyang He\textsuperscript{\rm 1,2}\textsuperscript{*} \qquad
Jie Wang\textsuperscript{\rm 2}\textsuperscript{*\#}  \qquad
Jiangning Zhang\textsuperscript{\rm 1}\textsuperscript{$\dagger$}  \qquad
Zhucun Xue\textsuperscript{\rm 1}  \qquad
}
\centerline{
Xingyuan Bu\textsuperscript{\rm 2}  \qquad
Qiangpeng Yang\textsuperscript{\rm 2}\textsuperscript{$\dagger$}  \qquad
Shilei Wen\textsuperscript{\rm 2}  \qquad
Lei Xie\textsuperscript{\rm 1}\textsuperscript{$\dagger$}  \qquad
} 
}
\affiliation[]{\textsuperscript{\rm 1} Zhejiang University \quad
\textsuperscript{\rm 2} ByteDance}
\contribution[*]{Equal contribution}
\contribution[\#]{Project Leader}
\contribution[\dagger]{Corresponding authors}

\abstract{
The quality and diversity of instruction-based image editing datasets are continuously increasing, yet large-scale, high-quality datasets for instruction-based video editing remain scarce. To address this gap, we introduce OpenVE-3M, an open-source, large-scale, and high-quality dataset for instruction-based video editing. It comprises two primary categories: spatially-aligned edits (Global Style, Background Change, Local Change, Local Remove, Local Add, and Subtitles Edit) and non-spatially-aligned edits (Camera Multi-Shot Edit and Creative Edit). All edit types are generated via a meticulously designed data pipeline with rigorous quality filtering. OpenVE-3M surpasses existing open-source datasets in terms of scale, diversity of edit types, instruction length, and overall quality. Furthermore, to address the lack of a unified benchmark in the field, we construct OpenVE-Bench, containing 431 video-edit pairs that cover a diverse range of editing tasks with three key metrics highly aligned with human judgment. We present OpenVE-Edit, a 5B model trained on our dataset that demonstrates remarkable efficiency and effectiveness by setting a new state-of-the-art on OpenVE-Bench, outperforming all prior open-source models including a 14B baseline.
}

\date{\today}

\checkdata[Project Page]{\url{https://lewandofskee.github.io/projects/OpenVE}}

\begin{document}
\maketitle

\input{sec/1_intro}
\input{sec/3_dataset}

\input{sec/4_method}
\input{sec/5_benchmark}
\input{sec/6_experiment}
\input{sec/7_conclusion}


\bibliographystyle{plainnat}
\bibliography{main}

\clearpage
\beginappendix
\input{sec/X_suppl}

\end{document}

%% file: sec/1_intro.tex
\vspace{-0.48cm}
\section{Introduction}
\label{sec:intro}
\begin{figure*}[t]
\centering
\includegraphics[width=1\linewidth]{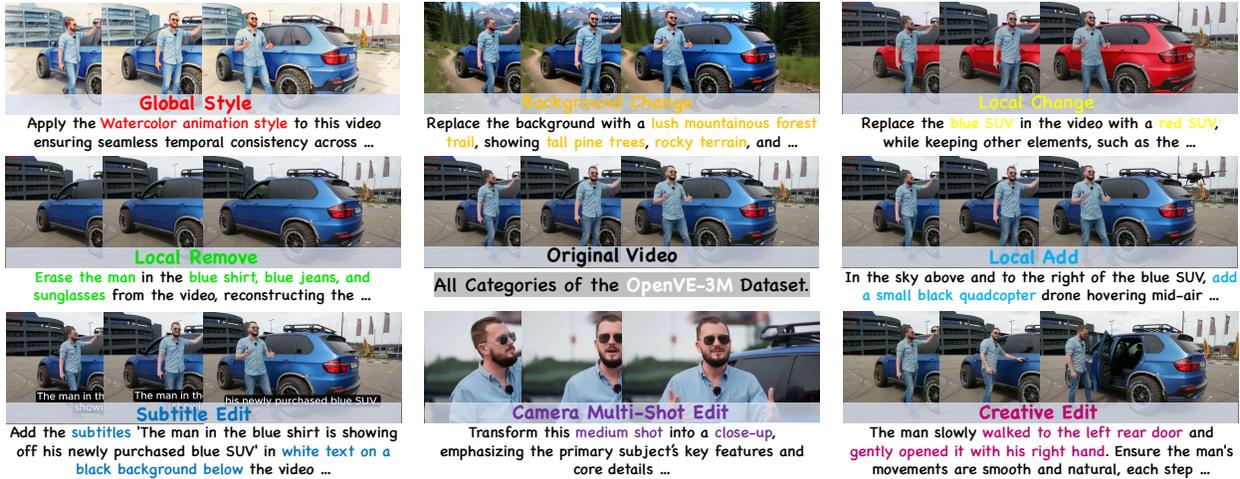}
\vspace{-0.7cm}
\caption{Demonstration of \textbf{Eight different categories} on the same video from the proposed \textbf{OpenVE-3M} dataset.}
\label{fig:datademo}
\vspace{-0.5cm}
\end{figure*}

Recently, \textbf{I}nstruction-guided \textbf{I}mage \textbf{E}diting (IIE) models such as FLUX-Kontext~\cite{kontext}, Qwen-Image-Edit~\cite{qwenimage}, and Nano Banana~\cite{gemini25} have demonstrated powerful editing capabilities. The success of an increasing number of data-driven, open-source image editing models can be attributed to the growing availability of high-quality, open-source, IIE datasets~\cite{anyedit,x2edit}. However, \textbf{I}nstruction-guided \textbf{V}ideo \textbf{E}diting (IVE) models and datasets remain \textit{under-explored}. This is mainly due to the significant difficulty in constructing high-quality IVE datasets. A key challenge is the need to maintain not only spatial consistency but also temporal consistency, which significantly increases the construction difficulty. Furthermore, the lack of effective data filtering strategies results in a high proportion of bad cases, making it difficult to completely eliminate poor-quality samples.

\begin{figure*}[t]
    \begin{minipage}[b]{0.63\textwidth}
        \centering
        \includegraphics[width=\linewidth]{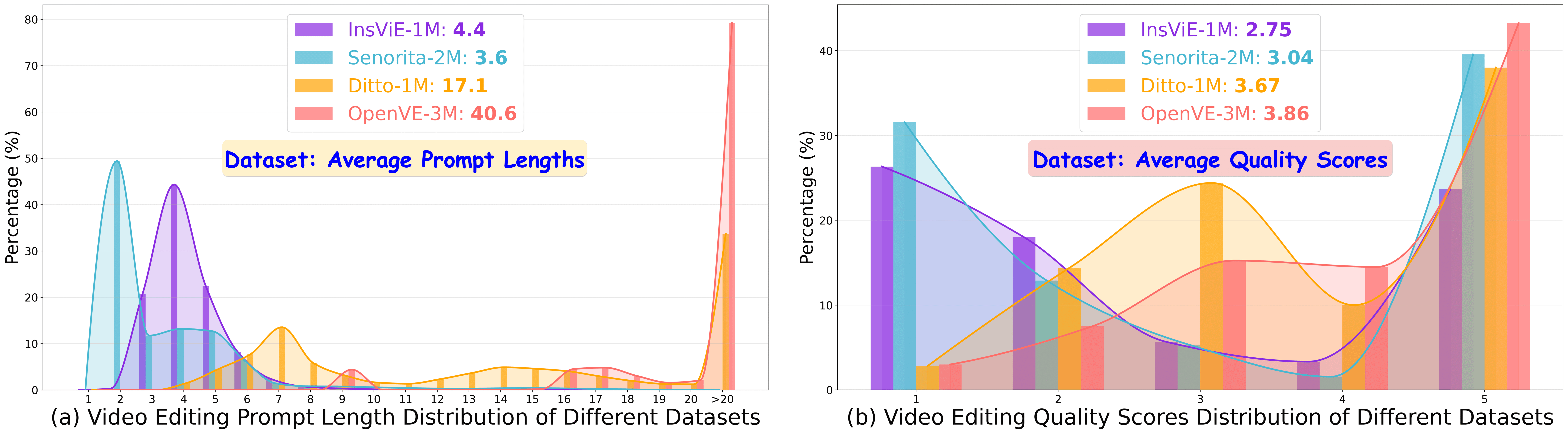}
        \caption{\textbf{Comparisons on video statistics} between OpenVE-3M and current open-source video editing datasets.}
        \label{fig:compdata}
    \end{minipage}%
    \hfill 
    \begin{minipage}[b]{0.36\textwidth}
        \centering
        \renewcommand{\arraystretch}{1.5}
        \resizebox{\linewidth}{!}{
            \begin{tabular}{cccccc}
            \toprule[0.1em]
            Datasets    & Amount & Cat. & Avg. Ins. Lgth & Frames & Resolution          \\
            \midrule
            EffiVED~\cite{effived}     & 155K   & 4       & N/A                      & 8      & 512×512             \\
            InsV2V~\cite{insv2v}      & 404K   & 4       & N/A                        & 16     & 256×256             \\
            VIVID-10M~\cite{vivid}   & 10M    & 3       & N/A                      & 30     & 1280×720            \\
            InsViE-1M~\cite{insvie}   & 1M     & 4       & 4.4                      & 25     & 1024×576            \\
            Señorita-2M~\cite{senorita} & 2M     & 6       & 3.6                      & 33-64  & \parbox[c]{2cm}{\centering 336×592 -\\ 1120×1984} \\
            Ditto-1M~\cite{ditto}    & 1M     & 3       & 17.1                     & 101    & 1280×720            \\
            \hline
            OpenVE-3M(Ours)        & 3M     & 8       & 40.6                     & 65-129 & 1280×720        \\
            \toprule[0.1em]
            \end{tabular}
        }
        \captionof{table}{Comparison with current IVE datasets. Cat./Avg. Ins. Lgth refers to Categories/Average Instruction Lengths respectively.}
        \label{tab:compdata}
    \end{minipage}
    \vspace{-1em}
\end{figure*}

Existing IVE datasets, such as InsViE-1M~\cite{insvie}, Senorita-2M~\cite{senorita}, and Ditto-1M~\cite{ditto}, suffer from four main issues: \textbf{small scale, limited edit types, short instructions, and poor editing quality}. \cref{tab:compdata} presents a quantitative analysis of existing open-source IVE datasets.
Although VIVID~\cite{vivid} contains 10M samples, it cannot be directly used for training as it only provides video masks instead of the edited videos. In contrast, InsViE-1M~\cite{insvie}, Senorita-2M~\cite{senorita}, and Ditto-1M~\cite{ditto} contain only 1-2M samples and offer a limited variety of editing types. \cref{fig:compdata}(a) shows the distribution of instruction lengths.
The average instruction length in~\cite{insvie} and~\cite{senorita} is only about 4 words, which is insufficient to provide precise editing guidance and thus negatively impacts performance. To assess dataset quality, we input the original video, edited video, and instruction into Gemini 2.5 Pro~\cite{gemini25} for evaluation on a 1-to-5 scale across three dimensions: \textit{Instruction Compliance}, \textit{Consistency \& Detail Fidelity}, and \textit{Visual Quality \& Stability}. For each dataset, we randomly sample 50 edited pairs from each category. The final score distributions are shown in \cref{fig:compdata}(b). Although ~\cite{insvie} and~\cite{senorita} show a notable number of 5-point scores, they also have a high proportion of 1-point bad cases, resulting in a low average quality score. \cite{ditto} demonstrates decent quality, but its edits are primarily focused on style transfer, lacking diversity. In summary, there is a clear need for a large-scale, high-quality, and diverse IVE dataset.

Therefore, we introduce \textbf{OpenVE-3M}, a large-scale, high-quality, and multi-category IVE dataset. It comprises \textbf{3M} samples categorized into two main types: \textbf{S}patially-\textbf{A}ligned (SA) and \textbf{N}on-\textbf{S}patially-\textbf{A}ligned (NSA). SA edits ensure that the edited video maintains motion consistency with the original video in both space and time, including six sub-types: \textit{Global Style}, \textit{Background Change}, \textit{Local Change}, \textit{Local Remove}, \textit{Local Add}, and \textit{Subtitles Edit}. NSA edits maintain the primary subject's consistency but introduce new, creative motion, including two sub-types: \textit{Camera Multi-Shot Edit} and \textit{Creative Edit}. Visual examples for all categories are presented in \cref{fig:datademo}.
Furthermore, OpenVE-3M features the longest average instruction length (40.6 words) and the highest average video editing quality score of 3.86 compared to existing IVE methods in Tab.~\ref{tab:compdata}.

The construction of the OpenVE-3M dataset consists of 3 stages, each leveraging many models and APIs for high-quality generation:
\textbf{\textit{1)}}\textbf{Data Pre-processing}: Preparing data for category-specific generation by employing models such as \textbf{M}ultimodal \textbf{L}arge \textbf{L}anguage \textbf{M}odel (MLLM)~\cite{qwen25vl}, detection and segmentation models~\cite{groundedsam2}, depth and edge estimation models~\cite{videodepthanything,opencv}, and local descriptors~\cite{describeanything}.
\textbf{\textit{2)}}\textbf{Taxonomy-guided Video Pair Generation}: Creating edited video pairs using a variety of models, including IIE model~\cite{kontext}, image-to-video model~\cite{wan} and so on.
\textbf{\textit{3)}}\textbf{High-Quality Pair Filtering}: Filtering the generated pairs to retain only high-quality examples using advanced MLLMs~\cite{gemini25,seed15vl,internvl35,qwen25vl}.
The detailed construction pipeline is described in \cref{sec:dataset}.

We also propose \textbf{OpenVE-Edit}, an IVE model composed of three main components: a MLLM, a \textbf{M}ixture-\textbf{o}f-\textbf{E}xperts (MoE) Connector, and a \textbf{Di}ffusion \textbf{T}ransformer (DiT). The MLLM processes both the video and the text instruction simultaneously. Unlike approaches that only use a text encoder for the instruction, our MLLM is designed to extract high-level instruction representations while also capturing spatio-temporal relationships within the video. We introduce an MoE-Connector which leverages different expert networks to process features corresponding to various editing types. To enhance training efficiency, we initialize the final linear layer of the experts to zero. This strategy prevents random features from disrupting the generation process in the early stages of training, thereby improving training stability and efficiency. The detailed model design is described in Sec.~\ref{sec:method}

Currently, there is no universal benchmark for IVE that is highly aligned with human evaluation. Therefore, we propose \textbf{OpenVE-Bench}, a meticulously curated benchmark comprising 431 video pairs across 8 categories. For each category, we design specific evaluation prompts targeting three key dimensions: \textit{Instruction Compliance}, \textit{Consistency \& Detail Fidelity}, and \textit{Visual Quality \& Stability}. The final score is obtained by feeding the original video, the edited video, and the instruction into a MLLM.

In summary, our contributions are as follows:
\begin{itemize}
    \item We introduce \textbf{OpenVE-3M}, a large-scale, high-quality, and diverse IVE dataset. It contains 3M samples across two major categories (SA and NSA) and 8 sub-categories.
    \item We propose a \textbf{Robust and Scalable Pipeline} for constructing high-quality, multi-category IVE data, aiming to facilitate further research in the community.
    \item We develop \textbf{OpenVE-Edit}, an efficient and effective IVE model. With only 5B parameters, it achieves SOTA performance, surpassing existing open-source 14B models.
    \item We establish \textbf{OpenVE-Bench}, a universal, multi-category, and challenging benchmark for IVE. It evaluates model performance along three key dimensions and demonstrates high alignment with human judgment.
\end{itemize}

\begin{figure}[t]
\centering
\includegraphics[width=1\linewidth]{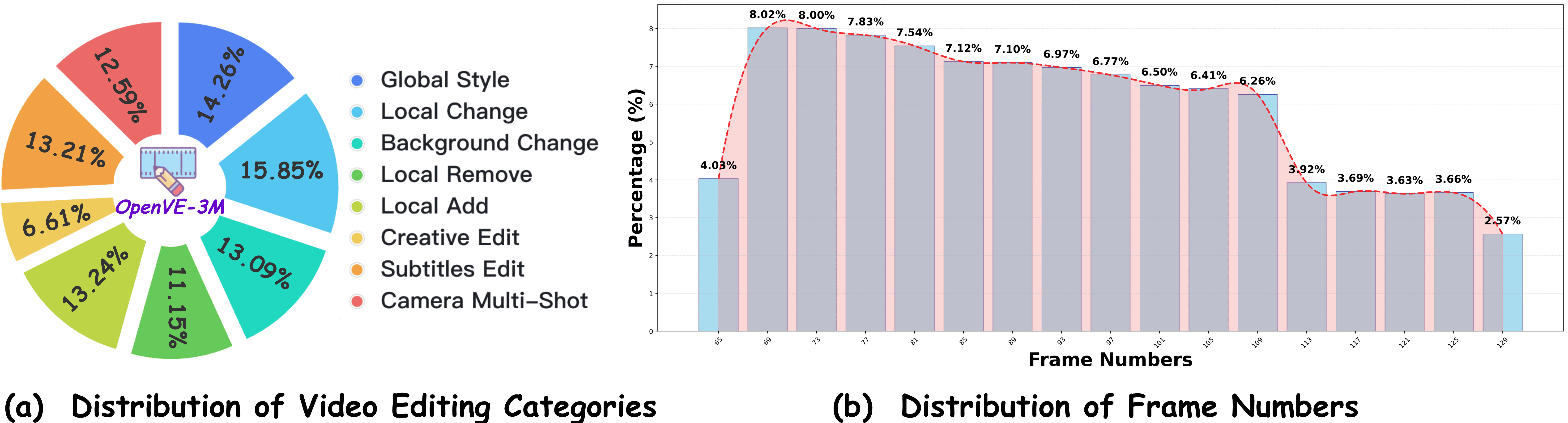} 
\caption{\textbf{Category} and \textbf{Frame Count} Statistics for OpenVE-3M.}
\label{fig:datastat}
\vspace{-0.5cm}
\end{figure}

%% file: sec/3_dataset.tex
\begin{figure*}[t]
\centering
\includegraphics[width=1\textwidth]{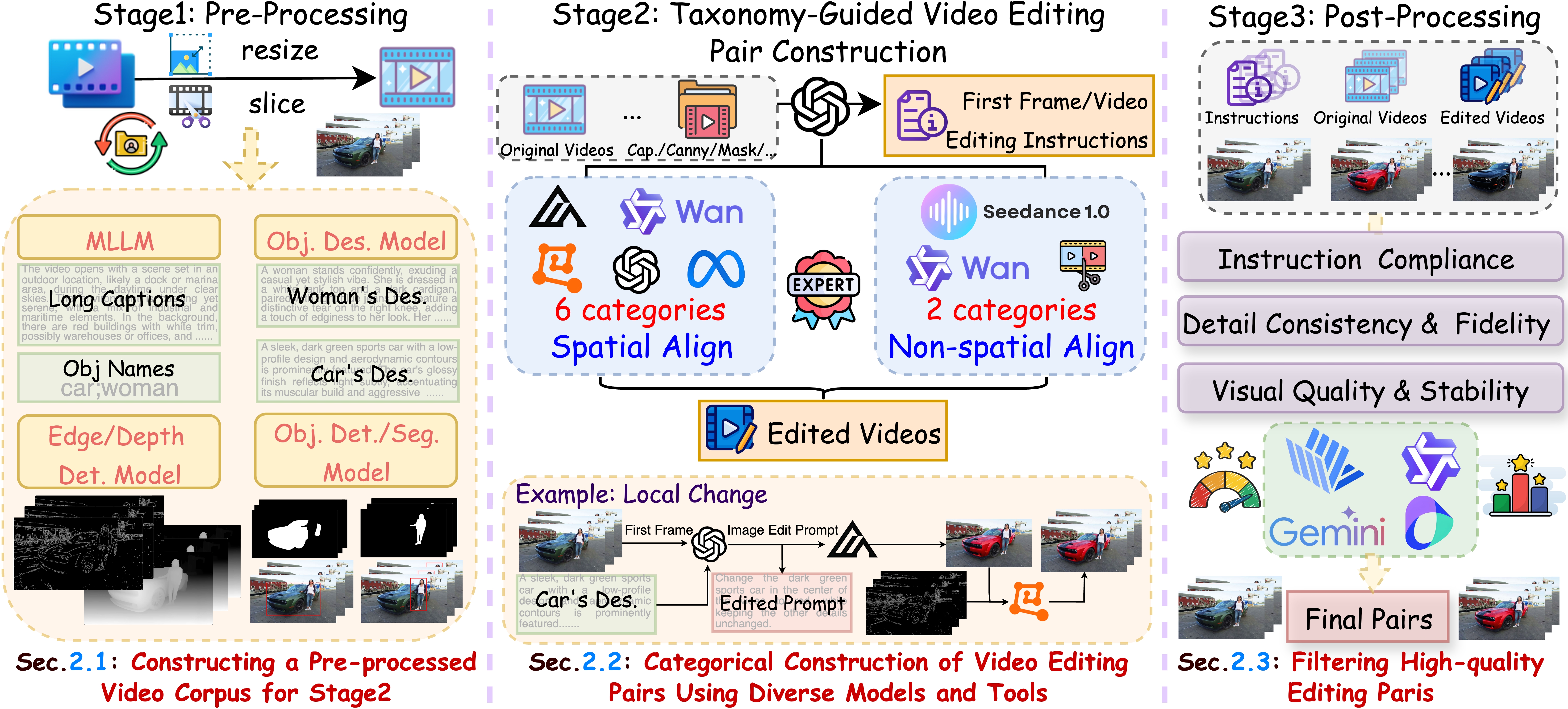} 
\caption{\textbf{Overview of Data Pipeline}.
Stage 1: Aims to construct a video corpus and perform various pre-processing steps in preparation for Stage 2 in Sec.~\ref{sec:sec21}.
Stage 2: Focuses on generating editing pairs for each category by leveraging a diverse set of models and tools in Sec.~\ref{sec:sec22}.
Stage 3: Involves the fine-grained filtering of all pairs generated in Stage 2 to retain only high-quality samples in Sec.~\ref{sec:sec23}.}
\label{fig:pipline}
\vspace{-1em}
\end{figure*}
\section{OpenVE-3M: A High-Quality Instruction-Guided Video Editing Dataset}
\label{sec:dataset}
We introduce OpenVE-3M, a large-scale, high-quality, multi-category, and balanced dataset designed for IVE. It comprises eight categories, divided into six \textbf{S}patially-\textbf{A}ligned~(SA) and two \textbf{N}on-\textbf{S}patially-\textbf{A}ligned~(NSA) types. The full construction pipeline is illustrated in Fig.\ref{fig:pipline}.
\vspace{-0.5em}
\subsection{Video Clip Pre-Processing}
\vspace{-0.5em}
\label{sec:sec21}
We establish a base video library of 1M videos sourced from three open-source high-quality datasets: Open-Sora-Plan~\cite{opensoraplan}, OpenViD-HD~\cite{openvid}, and UltraVideo~\cite{ultravideo}. For each source video, we randomly extract a clip of \textit{65 to 129 frames} and resize it to \textit{720P} resolution. Qwen2.5-VL-72B~\cite{qwen25vl} is used to generate a long-form description for the video clip while simultaneously extracting the names of detectable objects for subsequent processing. Concurrently, we generate depth and Canny edge maps for the clip using Video DepthAnything~\cite{videodepthanything} and OpenCV~\cite{opencv}, respectively. For the extracted object names, we employ Grounded SAM2~\cite{groundedsam2} to perform detection and segmentation, obtaining bounding boxes and mask videos for all identified objects. In parallel, the bounding boxes and the video clip are fed into the DAM~\cite{describeanything} to generate localized descriptions for each object. This process is applied to every source video, and the resulting information is used in the second stage to construct video editing pairs across different categories.
\vspace{-0.5em}
\subsection{Taxonomy-Guided Generation of Video Edits and Instructions}
\vspace{-0.5em}
\label{sec:sec22}
The OpenVE-3M dataset includes eight major video editing categories: six SA (\textit{Global Style Transfer, Background Change, Local Change, Local Remove, Local Add, Subtitles Edit}) and two NSA (\textit{Multi-Shot Camera Edit, Creative Edit}). SA edits maintain perfect consistency in motion and detail between the original and edited videos. In contrast, NSA edits maintain the primary subject’s consistency but introduce new, creative motion. The construction process for each category is detailed below. The respective flowcharts can be found in the \textbf{Appendix}.~\ref{sec:sup_ppl}.

\noindent \textbf{Global Style}. This category involves transforming the global style of a video while preserving the original motion and details. It includes \textit{18 common styles} (e.g., Ghibli, oil painting), \textit{four times of day} (e.g., Morning, Blue Hour), and \textit{three weather conditions} (e.g., Sunny, Rainy, Snowy). For the data generation pipeline, the initial frame of the video is input to GPT-4o with a carefully designed prompt to generate an image editing instruction and corresponding forward/backward prompts for video editing. The initial frame and the image editing instruction are then used with the FLUX-Kontext-dev~\cite{kontext} model to perform stylistic image editing. The edited initial frame, along with the Canny or depth video of the original clip, is then input to the Wan2.1-Fun-V1.1-14B-Control~\cite{wan} to generate the final stylized video under the structural guidance of the Canny maps.

\noindent \textbf{Background Change}. For videos with a clear foreground-background distinction, this task involves \textit{changing the background to various scenes}. Since robust video foreground detection algorithms are not readily available, we adapt an image-based approach. First, an image foreground detection algorithm is applied to the initial frame to generate a foreground mask. We then compute the Intersection over Union (IoU) between this mask and all possible combinations of object masks (obtained via Grounded SAM2~\cite{groundedsam2}). If the IoU for a combination exceeds 0.95, that combined mask is considered the definitive foreground mask for the video. The pipeline then proceeds by using GPT-4o to generate image and video-level prompts for background editing. The FLUX-Kontext-dev~\cite{kontext} model edits the background of the initial frame. We create a foreground-only Canny/depth video by masking out the background. Finally, this masked Canny/depth video and the edited initial frame are input to the Wan2.1-Fun-V1.1-14B-Control~\cite{wan} model to generate the final video with the new background.

\noindent \textbf{Local Change}. This includes a range of edits such as \textit{object transformation, style modification, color alteration, and age progression}. The initial frame and its localized object descriptions (from DAM~\cite{describeanything}) are input to GPT-4o with a carefully designed prompt. This generates a variety of rich, localized image and video editing prompts. The FLUX-Kontext-dev~\cite{kontext} model then performs the local edit on the initial frame. Finally, the edited frame, along with the original Canny or depth video, is fed into the Wan2.1-Fun-V1.1-14B-Control~\cite{wan} model to produce the edited video clip.

\noindent \textbf{Local Remove/Add}. For local object removal and addition, we devise two distinct data generation pipelines. To create training data for local addition, we first leverage video inpainting models~\cite{diffueraser,minimaxremover} to remove an object from an original video clip based on a given mask. This resulting inpainted video, where the object is absent, serves as the source video for the addition task. We then use GPT-4o to generate a corresponding `local add' instruction from an object description~\cite{describeanything}, while the original, unedited video acts as the ground-truth target. Conversely, to generate data for local removal, we synthesize a video with a new object. This process begins by using GPT-4o to generate an image-level `add' instruction for the first frame of a video, which is then edited by the FLUX-Kontext-dev~\cite{kontext} model. Subsequently, the Wan2.2-I2V-A14B~\cite{wan} model generates a full video from this single edited frame, and we employ Grounded SAM2~\cite{groundedsam2} to segment the newly added object. The masked region containing the object is then pasted onto the original video frames to produce the final edited clip. This synthetic video with the added object then becomes the source video for the removal task. A `local remove' instruction is generated by GPT-4o, and the original video without object serves as the corresponding target video.

\noindent \textbf{Subtitles Edit}. This category includes tasks for \textit{adding, removing, and modifying subtitles}, featuring \textit{nine variations} (\textit{three positions: top, middle, bottom}). The pipeline uses the GPT-4o to generate appropriate text for the video. The rendering tool is then used to render subtitles of different styles at various positions. Finally, GPT-4o generates the editing instructions for the add, remove, and modify tasks.

\noindent \textbf{Camera Multi-Shot Edit}. This NSA task involves editing a video to switch between \textit{close-up, medium, and wide shots of the same subject}, comprising a total of \textit{six transition types}. We leverage the ``Native Multi-Shot'' feature of Seedance~\cite{seedance}. GPT-4o generates prompts for a three-shot sequence, which are then used for I2V generation. After generating a multi-shot video with a consistent subject, visual style, and atmosphere, a shot detection model segments the different shots. These segments are then used to create shot-switching edit pairs and corresponding instructions.

\noindent \textbf{Creative Edit}. This NSA involves editing an object to follow a creative instruction, where the \textit{subject's actions may change significantly}. For a given source video, we use GPT-4o to generate multiple creative I2V instructions based on its initial frame. The Seedance~\cite{seedance} model then generates high-quality videos for each instruction. Finally, any two of these generated videos can form an editing pair, with GPT-4o generating a creative editing instruction to describe the transformation between them.



\begin{figure*}[t]
\centering
\includegraphics[width=1\textwidth]{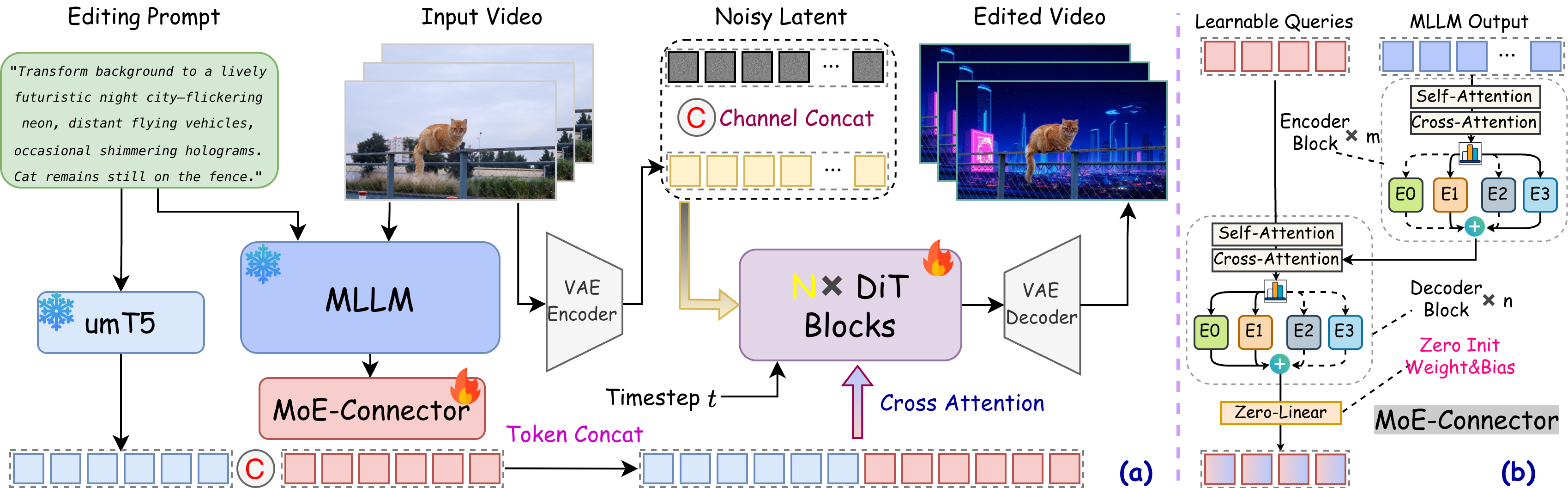} 
\caption{\textbf{Overall of the OpenVE-Edit}. (a) Architecture of OpenVE-Edit. (b) The detailed structure of the MoE-Connector module.}
\label{fig:model}
\vspace{-1.5em}
\end{figure*}
\vspace{-0.5em}
\subsection{Data Filtering and Curation}
\vspace{-0.5em}
\label{sec:sec23}
We designed a meticulous data filtering pipeline for all synthesized data pairs (excluding Subtitles Edit). This process is based on three main evaluation metrics, each rated on a scale of 1 to 5: 1) \textit{Instruction  Compliance}, 2) \textit{Consistency \& Detail Fidelity}, and 3) \textit{Visual Quality \& Stability}. Crucially, the score for Instruction Compliance serves as an upper bound for the other two metrics. We prioritize instruction adherence as the primary evaluation criterion because many generated videos, despite high visual quality, may fail to follow the edit instruction.

We then input the editing instruction, original video, and edited video into various VLMs for automated scoring. To benchmark the VLMs, we first manually annotated a set of 300 video pairs, defining pairs with an average score $>3$ as positive samples and those $\leq3$ as negative samples. The resulting accuracies were: Qwen3-VL-30B-A3B~\cite{qwen25vl} (61\%), InternVL3.5-38B~\cite{internvl35} (66\%), Seed1.6-VL~\cite{seed15vl} (70\%), and Gemini2.5-Pro~\cite{gemini25} (69\%). Due to API rate limits (TPM) for Seed1.6-VL and Gemini2.5-Pro, we ultimately selected the Intern3.5-VL-38B~\cite{internvl35} model to score the entire dataset, retaining all pairs with a score $\geq3$.

Our final OpenVE-3M dataset contains 3 million video editing pairs across the eight categories. As shown in Tab. ~\ref{tab:compdata}, OpenVE-3M surpasses existing IVE datasets in terms of task diversity, data scale, resolution, frame count, and prompt detail. Specifically, our dataset features more categories and a higher average instruction length than its counterparts. Its video frame count and resolution are also among the highest quality available. Furthermore, as illustrated in Fig.~\ref{fig:datastat}, the distribution across different editing types is well-balanced, making OpenVE-3M a comprehensive, high-quality, and large-scale IVE dataset.

%% file: sec/4_method.tex
\section{OpenVE-Edit: An Effective Instruction-Guided Video Editing Method}
\label{sec:method}
In Fig.~\ref{fig:model}, OpenVE-Edit, consists of three main modules: a MLLM, a MoE-Connector, and a DiT. The input editing instruction and video are jointly fed into the MLLM to capture the semantic relationships between the instruction and the visual content. Subsequently, a task-aware MoE-Connector processes the hidden features from the MLLM, decoupling them through multiple expert networks. These processed features are then concatenated along the token dimension with the instruction features encoded by umT5. Concurrently, the latent features of the original video, derived from a VAE, are concatenated with noise along channel dimension. This composite latent representation subsequently interacts with the combined semantic editing features through the Cross-Attention mechanism in the DiT model.
\vspace{-0.5em}
\subsection{Editing Knowledge Injection via MLLM}
\vspace{-0.5em}
Baseline video generation models, such as Wan~\cite{wan}, rely solely on umT5-encoded text features, which interact with noisy latent features through cross-attention. However, using only umT5 features provides a literal representation of the editing instruction and fails to capture higher-level semantic and spatial relationships between the instruction and the visual content. Therefore, we jointly input the original video and the editing instruction into an MLLM. This enables the model to capture these higher-level semantic and spatial relationships. Furthermore, to emphasize edit-relevant semantics, we discard token embeddings associated with prefixes, such as system prompts. This filtering process ensures that only tokens corresponding directly to the editing information are retained, focusing the subsequent processing on the specific editing task.
\subsection{Decoupling Task Cues from Expert Gating}
Addressing the diverse requirements of editing tasks with a single model presents a challenge of task heterogeneity, which can lead to parameter inefficiency. Shared parameters may internalize conflicting representations from different tasks, resulting in suboptimal specialization and an increased parameter count. Therefore, we design a task-aware MoE-Connector module to handle different editing types for both images and videos. Since videos of varying frame counts and resolutions produce different numbers of tokens from the MLLM, we employ learnable queries to extract key information from the hidden states for different editing tasks. This is achieved by activating specific sparse expert sub-networks, allowing for precise capacity allocation for each task.
Given input $\mathbf{X}_{in} \in \mathbb{R}^{b \times s \times d}$ where $b$ is batch size, $s$ is sequence length, and $d$ is the MLLM hidden dimension, the forward process of the overall model is:
\vspace{-0.2cm}
\begin{equation}
\vspace{-0.2cm}
\mathbf{Y} = \mathbf{W}_{o}\Big(\text{MoE-Dec}\big(\mathbf{Q}_{l}, \text{MoE-Enc}(\mathbf{F}_{i}(\mathbf{X}_{in}))\big)\Big),
\end{equation}
where $\mathbf{Q}_{l} \in \mathbb{R}^{b \times L_q \times D_h}$ is the learnable decoder query with $L_q$ being the learnable query length and $D_h$ being the hidden dimension, $\mathbf{W}_{o}$ is the output linear transformation matrix, $\mathbf{F}_{i}$ is the Feed-Forward Network~(FFN) for dimension mapping, $\text{MoE-Enc}$ and $\text{MoE-Dec}$ are the MoE encoder and decoder modules respectively, each containing a self-attention layer, a cross-attention layer, and an MoE-FFN.
For input $\mathbf{x} \in \mathbb{R}^{D_h}$, the MoE-FFN computation process is:
\vspace{-0.2cm}
\begin{equation}
\vspace{-0.2cm}
\mathbf{y} = \sum_{i \in S} w_i \cdot \Big(\mathbf{W}_{i,2} \cdot \text{GELU}(\mathbf{W}_{i,1} \mathbf{x} + \mathbf{b}_{i,1}) + \mathbf{b}_{i,2}\Big),
\end{equation}
where $S = \text{topk}(\text{softmax}(\mathbf{W}_g \mathbf{x}), k)$ is the set of top-$k$ selected expert indices, $\mathbf{W}_g$ is the gate weight matrix, $\mathbf{W}_{i,j}$ and $\mathbf{b}_{i,j}$ are the weights and biases of the $i$-th expert network, $w_i$ is the normalized weight for expert $i$.
\vspace{-0.5em}
\subsection{Accelerating Convergence in A Unified Model}
\vspace{-0.5em}
Unlike large pre-trained models, our MoE-Connector is randomly initialized, which risks introducing noise and destabilizing training. To mitigate this, we draw inspiration from ControlNet~\cite{controlnet} by \textit{zero-initializing} the weights of the connector's final MLP layer $\mathbf{W}_{o}$. We then concatenate its output with the original umT5 instruction features. This design effectively renders the module "invisible" at the start of training, as its zero-output preserves the integrity of the umT5 features. As training progresses, the module gradually learns to contribute useful information via gradient descent, ensuring stable and efficient convergence.

%% file: sec/5_benchmark.tex
\section{OpenVE-Bench: A Robust Instruction-Guided Video Editing Benchmark}
\label{sec:bench}
\subsection{Benchmark Construction}
\vspace{-0.5em}
Our {OpenVE-Bench} is constructed with two primary categories: SA and NSA edits. These are further divided into eight fine-grained subcategories, totaling 431 IVE pairs, with each subcategory containing over 50 video clips on average. For {\textit{Global Style}}, we carefully select 58 video clips and design instructions covering 18 distinct styles, 3 different times of day, and 4 weather conditions. For {\textit{Background Change}}, we manually select 59 clips suitable for background changes and design diverse instructions with styles ranging from harmonious to highly stylized. For {\textit{Local Change}}, we select 65 clips featuring prominent subjects such as humans, animals, or vehicles, with instructions for object transformation, style modification, and age manipulation. For {\textit{Local Add}}, we select 67 clips of varying difficulty and design instructions for adding objects of different sizes, from small items like a kite to large ones like a car. For {\textit{Local Remove}}, we select 59 videos featuring subjects of varying sizes and types and design corresponding removal instructions. For {\textit{Subtitles Edit}}, we select 50 clips from diverse scenes and styles, with instructions to add, delete, or replace text at top, middle, or bottom positions using various styles. For {\textit{Camera Multi-Shot Edit}}, we select 43 clips with prominent subjects and design instructions for transitioning between long, medium, and close-up shots. For {\textit{Creative Edit}}, we select 30 distinctive video clips and use GPT-4o to generate diverse and imaginative instructions.
\subsection{Evaluation Metrics}
Following established criteria~\cite{imgedit,step1xedit,complextedit}, we evaluate IVE on a 1--5 scale for \textit{Instruction Compliance}, \textit{Consistency \& Fidelity}, and \textit{Visual Quality \& Stability}. Crucially, the \textit{Instruction Compliance} score upper-bounds the other two, prioritizing adherence to the edit instruction. We also assess video quality by quantifying temporal flicker with \textit{TF} (Temporal Flickering, measured by mean absolute difference) and consistency with \textit{CLIP-F} (average inter-frame CLIP similarity).

%% file: sec/6_experiment.tex
\section{Experiments} \label{sec:exp}
\subsection{Implementation Details}
We use Wan2.2-TI2V-5B as the video generation base model. The Qwen2.5VL-3B~\cite{qwen25vl} processes the input video and editing instructions. For the MoE-Connector, both the encoder and decoder blocks have 2 layers. Each layer contains 6 experts, with 2 experts activated per forward pass. The sequence length of the learnable queries is 512. We employ a two-stage training strategy. First, we train the model for one epoch at a 480P resolution with a global batch size of 512 and a learning rate of 1e-5. Subsequently, we fine-tune the model for an additional epoch at a 720P resolution with a reduced learning rate of 1e-6.
\subsection{Comparison With State-of-The-Art Models}
We compare our model with existing state-of-the-art open-source models, including VACE~\cite{vace}, OmniVideo~\cite{omnivideo}, InsViE~\cite{insvie}, AnyV2V~\cite{anyv2v}, Senorita~\cite{senorita}, ICVE~\cite{icve}, Lucy-Edit~\cite{lucyedit}, DITTO~\cite{ditto}, and VideoCoF~\cite{videocof}, as well as the closed-source model Runway Aleph~\cite{runway}. During our reproduction of the open-source models on a single GPU with 80GB of VRAM, we observe several limitations. OmniVideo~\cite{omnivideo} can only generate videos at a 640x352 resolution with 17 frames; other settings result in video artifacts. The ICVE~\cite{icve} model can generate a maximum of 41 frames at a 480x768 resolution; generating more frames leads to OOM. Therefore, we use a resolution of 384x240 for ICVE to ensure all frames can be edited.
Unlike end-to-end approaches, AnyV2V~\cite{anyv2v} and Senorita~\cite{senorita} rely on external I2I models to edit the first frame before guiding the final video generation. AnyV2V employs InstructPix2Pix~\cite{instructpix2pix}, while Senorita-2M uses FLUX-Kontext~\cite{kontext} for this first-frame editing.
The other models are evaluated using their respective official training resolutions and input frame counts. Additionally, due to the cost constraints of Runway Aleph, we only select 30 samples from each evaluation category for benchmarking.

\begin{table*}[t]
    \centering
    \renewcommand{\arraystretch}{1.0}
    \setlength\tabcolsep{5.0pt}
    \caption{Quantitative comparison on \textbf{OpenVE-Bench} with Seed1.6-VL~\cite{seed15vl}. \protect\sethlcolor{oran_tab}\hl{Orange}, \protect\sethlcolor{gray_tab}\hl{gray}, and \protect\sethlcolor{blue_tab}\hl{blue} backgrounds indicate closed-source, open-source, and our models, respectively. \#Params. and \#Reso. refer to parameter counts and resolutions, respectively. * indicates that an extra image editing model is required for video editing.}
    \resizebox{1\linewidth}{!}{
        \begin{tabular}{c|cc|c|cccccccc|c|c|c}
        \toprule[0.1em]
\multirow{2}{*}{Methods} & \multicolumn{2}{c|}{Model Details} & \multicolumn{9}{c|}{OpenVE-Bench results on Seed1.6-VL} & \multirow{2}{*}{\makecell{Gemini \\ Overall}} & \multicolumn{2}{c}{Video Quality} \\
\cmidrule(lr){2-3} \cmidrule(lr){4-12} \cmidrule(lr){14-15}
& \#Params. & \#Reso. & Overall & GS & BC & LC & LR & LA & SE & CE & CME & & TF & CLIP-F \\
 \midrule
 \rowcolor{oran_tab}Runway~\cite{runway} & - & 1280$\times$720 & 3.50 & 3.47 & 2.84 & 3.88 & 3.88 & 2.79 & 3.50 & 3.23 & 4.48 & 3.65 & 0.9851 & 0.9659 \\
\rowcolor{gray_tab}VACE~\cite{vace} & 14B & 1280$\times$720 & 1.17 & 1.41 & 1.16 & 1.43 & 1.00 & 1.05 & 1.02 & 1.13 & 1.16 & 1.57 & 0.9852 & 0.9685 \\
\rowcolor{gray_tab}OmniVideo~\cite{omnivideo} & 1.3B & 640$\times$352 & 1.02 & 1.02 & 1.00 & 1.00 & 1.00 & 1.00 & 1.16 & 1.00 & 1.00 & 1.31 & 0.9916 & 0.9785 \\
\rowcolor{gray_tab}InsViE~\cite{insvie} & 2B & 720$\times$480 & 1.40 & 2.25 & 1.23 & 1.60 & 1.00 & 1.23 & 1.22 & 1.68 & 1.02 & 1.53 & 0.9768 & 0.9739 \\
\rowcolor{gray_tab}AnyV2V~\cite{anyv2v}* & 3.7B & 512$\times$512 & 1.41 & 1.84 & 1.31 & 1.94 & 1.09 & 1.27 & 1.02 & 1.57 & 1.11 & 1.51 & 0.9614 & 0.9589 \\
\rowcolor{gray_tab}Señorita~\cite{senorita}* & 5B & 800$\times$448 & 1.90 & 2.29 & 1.47 & 2.69 & 1.93 & 1.64 & 2.49 & 1.91 & 1.02 & 1.93 & 0.9802 & 0.9658 \\
\rowcolor{gray_tab}Lucy-Edit~\cite{lucyedit} & 5B & 1280$\times$704 & 1.95 & 2.17 & 2.20 & 3.30 & 1.03 & 2.37 & 1.06 & 2.35 & 1.14 & 2.15 & 0.9836 & 0.9692 \\
\rowcolor{gray_tab}ICVE~\cite{icve} & 13B & 384$\times$240 & 2.25 & 2.35 & 1.86 & 2.91 & 2.68 & 2.27 & 2.04 & 1.94 & 1.38 & 2.07 & 0.9877 & 0.9661 \\
\rowcolor{gray_tab}DITTO~\cite{ditto} & 14B & 832$\times$480 & 2.06 & 3.70 & 2.23 & 2.28 & 1.00 & 2.08 & 1.01 & 2.61 & 1.51 & 1.98 & 0.9848 & 0.9635 \\
\rowcolor{gray_tab}VideoCoF~\cite{videocof} & 14B & 1280$\times$704 & 2.24 & 1.07 & 1.33 & 3.52 & 3.26 & 2.67 & 2.57 & 1.64 & 1.05 & 2.24 & 0.9768 & 0.9651 \\
\midrule
\rowcolor{blue_tab}OpenVE-Edit & 5B & 1280$\times$704 & \textbf{2.41} & 3.11 & 2.72 & 3.19 & 1.42 & 2.41 & 2.56 & 2.01 & 1.24 & \textbf{2.49} & 0.9889 & 0.9698 \\
        \toprule[0.1em]
        \end{tabular}
    }
    \label{tab:result1}
\end{table*}
\noindent\textbf{Quantitative Comparison.}
Tab.~\ref{tab:result1} presents the evaluation results of all instruction-guided video editing models on OpenVE-Bench. The closed-source model, Runway Aleph, achieves the best performance on both the Seed-1.6VL~\cite{seed15vl} and Gemini 2.5 Pro~\cite{gemini25} evaluators, significantly outperforming existing open-source models. The open-source models VACE~\cite{vace}, OmniVideo~\cite{omnivideo}, and InsViE~\cite{insvie} exhibit limited performance due to constraints in their model size or training data. With 5B parameters, Lucy-Edit~\cite{lucyedit} achieves moderate performance. ICVE~\cite{icve}, with 13B parameters, obtains respectable results, but its high-resolution editing capability is limited to a smaller number of frames. Ditto~\cite{ditto} scores highly on the global style metric, as its training dataset primarily consists of this edit type. Our model, OpenVE-Edit, with only 5B parameters, achieves an overall score of \textbf{2.41}, surpassing all existing open-source models, including larger 13B and 14B baselines.
While OmniVideo~\cite{omnivideo} excels on temporal metrics (TF and CLIP-F), it entirely disregards the edit instruction. Our method, however, achieves both strong temporal quality and superior instruction-following results.

\begin{figure*}[t]
\centering
\includegraphics[width=1.0\linewidth]{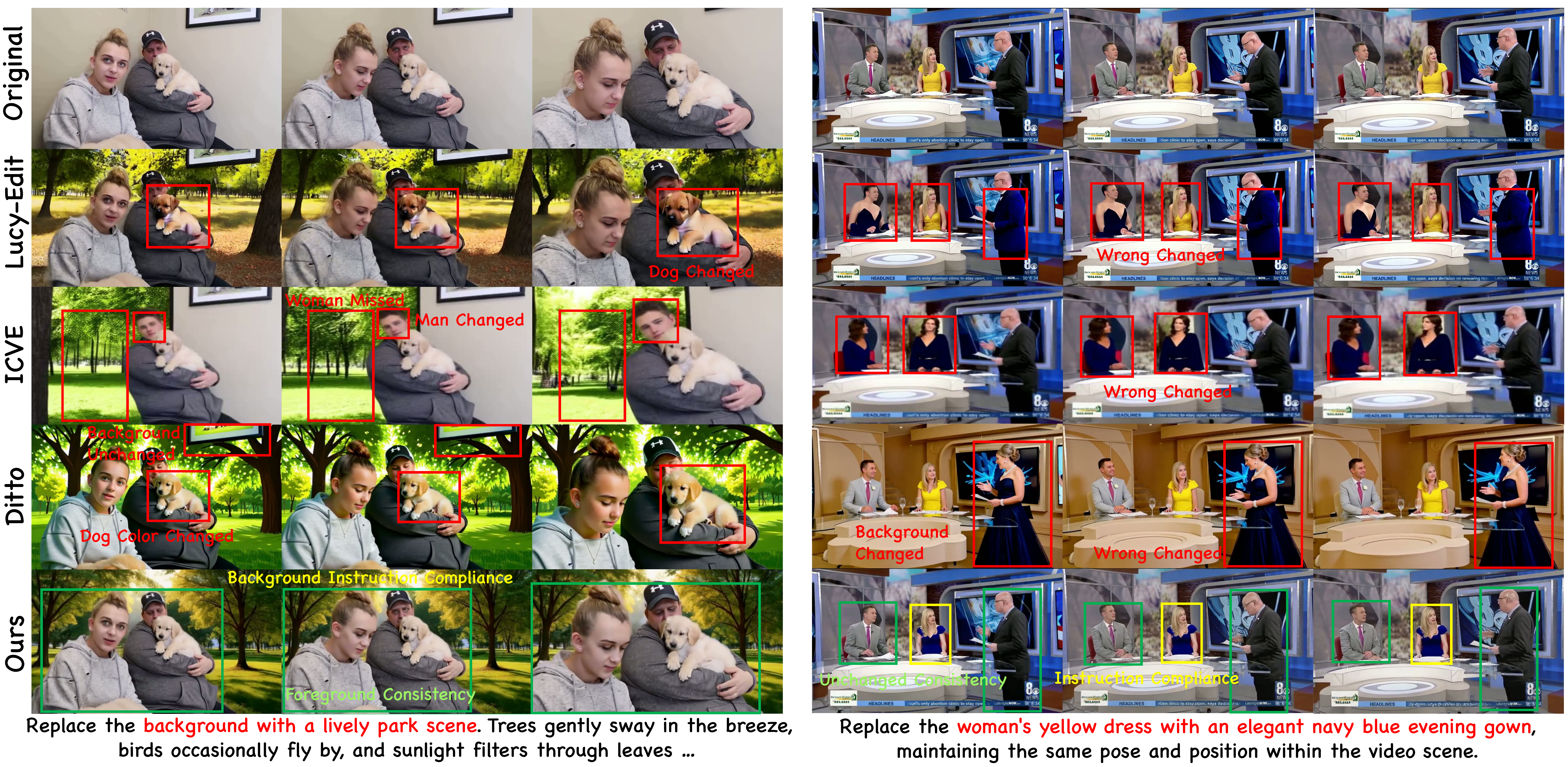}
\caption{Qualitative comparison with SoTA methods on Background Change (left) and Local Change (right) examples.}
\label{fig:qualitative_demo}
\end{figure*}

\noindent\textbf{Qualitative Results.}
Fig.~\ref{fig:qualitative_demo} presents a qualitative comparison between our method and existing open-source SOTA methods. We select the top three performing open-source models for this comparison. In the ``Background Change'' example on the left, although Lucy-Edit~\cite{lucyedit} successfully changes the background, it fails to maintain the consistency of the dog from the original video. ICVE~\cite{icve} incorrectly erases the woman and alters the man's appearance. Ditto~\cite{ditto} mistakenly treats the painting on the wall as a foreground object and darkens the dog's fur. In contrast, our method alters the background according to the instruction while preserving the consistency of all foreground subjects. In the ``Local Change'' example on the right, Lucy-Edit~\cite{lucyedit} incorrectly edits the clothing of all three individuals. ICVE~\cite{icve} erroneously modifies the two people on the left and also changes their identities. Ditto~\cite{ditto} not only edits the wrong subject but also incorrectly alters the background. Our method, however, precisely follows the instruction by changing only the specified woman's clothing while maintaining the consistency of the other individuals and the background. Additional qualitative results are provided in Appendix~\ref{sec:sup_model}.

\subsection{Extra Ablation and Explanatory Analysis}
\begin{figure}[b]
\centering
\includegraphics[width=1\linewidth]{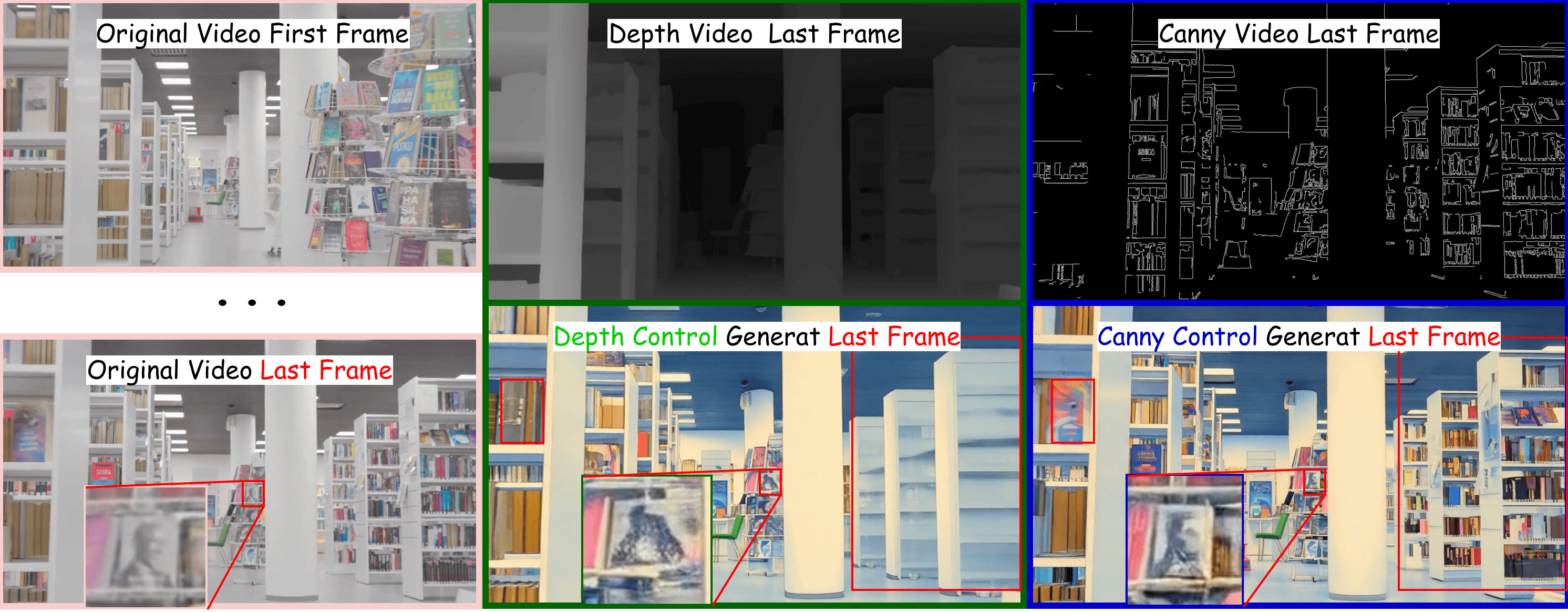} 
\caption{Comparison with different control videos for Stage2.}
\label{fig:ablation1}
\end{figure}
\noindent \textbf{Ablation on Control Signals for Data Construction.} We use depth maps and Canny edge maps derived from the original video as separate control conditions using Wan2.1-Fun-V1.1-14B-Control~\cite{wan}. In Fig.~\ref{fig:ablation1}, when there is a significant difference between the first and last frames of a video, using depth control results in poor detail generation. In contrast, Canny edge control better preserves the details of the original video. Furthermore, if the depth maps are consistent across consecutive frames, the background in the generated video often remains static. Therefore, we choose Canny maps as the control condition for our pipeline.

\noindent \textbf{Acceleration of Data Construction.} In the data construction phase, three tasks require inference with the Wan2.1-Fun-V1.1-14B-Control model. The average inference time for a single 81-frame, 720P video on one GPU is \textit{50 minutes}. This speed is too slow for large-scale data generation. Consequently, we employ two strategies to accelerate this process. First, we replace FlashAttention-2~\cite{flashattention2} with the faster SageAttention-2~\cite{sageattention2}. Second, we reduce the number of inference steps from \textit{50 to 10}. These optimizations reduce the average generation time to \textit{6 minutes}, an $\times8\uparrow$ speedup. Manual inspection of a large number of samples confirms that the quality of the accelerated data generation meets our requirements and shows no degradation. Additionally, for image-to-video (I2V) generation, we replace the original Wan2.2-I2V-A14B model with its \textit{4-step distilled} version, Wan2.2-I2V-A14B-NFE4-V1, which also significantly improves efficiency.

\begin{table}[t]
  \centering
  \renewcommand{\arraystretch}{0.7}
  \setlength\tabcolsep{5.0pt}
  \caption{Ablations of the \textbf{Model Design} and the \textbf{Training Dataset}. \textbf{Overall} indicates the average value of all metrics in OpenVE-Bench.}
  \resizebox{0.8\linewidth}{!}{
    \begin{tabular}{cccccc}
    \toprule
\multirow{2}{*}{Index} & \multicolumn{2}{c}{Structure} & \multicolumn{2}{c}{Training Dataset} & \multirow{2}{*}{Overall} \\ 
\cmidrule(r){2-3} 
\cmidrule(l){4-5}  
& MLLM     & MoE-Connector     &  Senorita-2M        & OpenVE-3M       &                          \\
\midrule
\circled{1} &         &                   &                    & $\checkmark$               & 2.12                     \\
\circled{2} & $\checkmark$        &                   &                    & $\checkmark$               & 2.31                      \\
\circled{3} & $\checkmark$        & $\checkmark$                 & $\checkmark$                  &                 & 1.54                     \\
\circled{4} & $\checkmark$        & $\checkmark$                 &                    & $\checkmark$               & 2.41      \\
\bottomrule 
\end{tabular}}
\label{tab:ablations}%
\end{table}
\begin{figure}[t]
\centering
\includegraphics[width=1\linewidth]{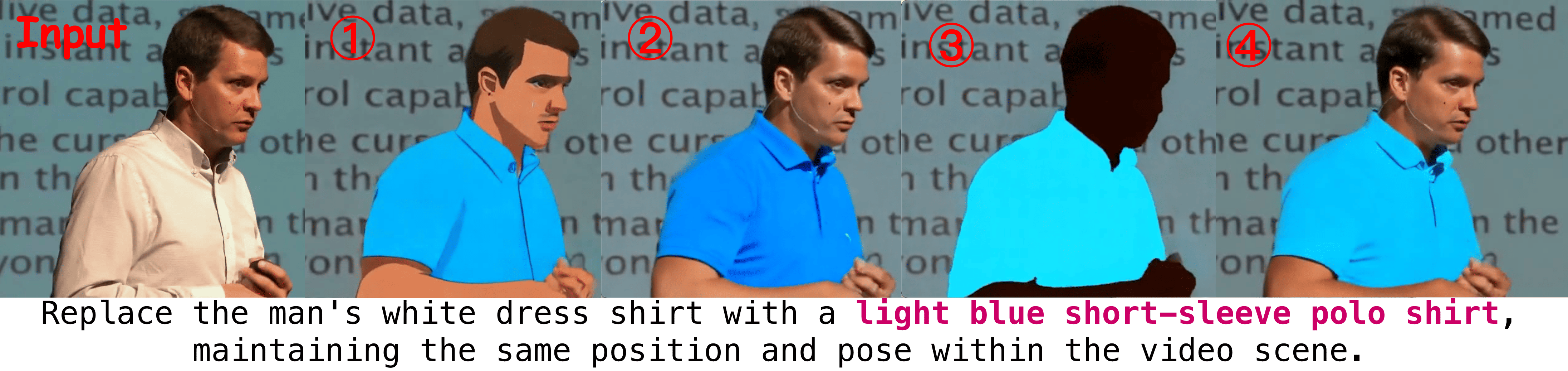}
\caption{Qualitative ablations of the \textbf{Model Design} and the \textbf{Training Dataset}.}
\label{fig:ablation2}
\end{figure}
\noindent \textbf{Ablations on Model Design.} Compared to using only umT5 instructions, jointly inputting the video and instruction into the MLLM enables the model to better comprehend the instruction and execute the corresponding edit. In Fig.~\ref{fig:ablation2}, \circled{1} using only T5 features can lead to a mismatch between the editing result and the instruction, whereas features from the MLLM \circled{2} mitigate this issue. This is corroborated by the quantitative metrics, with the score improving from \textit{2.12 to 2.31} ($+0.19\uparrow$) in Tab.~\ref{tab:ablations}. Furthermore, compared to a simple MLP connector with shared parameters, the MoE-Connector is more effective at processing features for different edit types by routing them to specialized experts. This is reflected in the average score, which increases from \textit{2.31 to 2.41} ($+0.1\uparrow$), demonstrating the effectiveness of our approach in handling diverse and complex editing tasks. \circled{4} enhances \circled{2} by incorporating an MoE-Connector, resulting in a better representation of semantic details.

\noindent \textbf{Ablations on Training Datasets.} We ablate the training data using Senorita-2M~\cite{senorita} as a baseline. The resulting average score of \textit{1.54} is significantly lower than the \textit{2.41} achieved when training on our OpenVE-3M dataset in Tab.~\ref{tab:ablations}. This indicates potential quality issues in the Senorita-2M dataset, a conclusion further supported by the high proportion of samples with a score of 1, as illustrated in Fig.~\ref{fig:compdata}. Also, in Fig.~\ref{fig:ablation2}, the model in \circled{4} is outperformed by \circled{3}, a performance drop caused by training on the low-level data abundant in Senorita-2M.

\begin{figure}[b]
\centering
\includegraphics[width=1\linewidth]{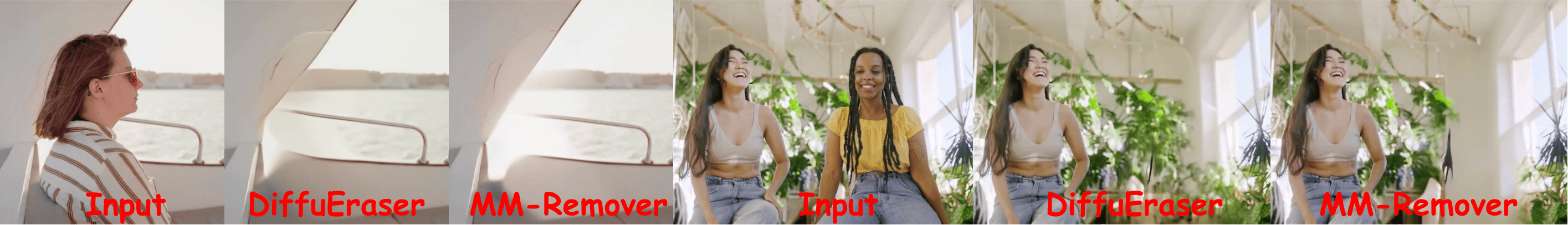}
\caption{Qualitative ablations of the \textbf{Video Removal Methods}.}
\label{fig:ablation3}
\end{figure}

\noindent \textbf{Ablations on Video Removal Methods.} While MiniMax-Remover~\cite{minimaxremover} offers smoother temporal consistency and less flicker than DiffuEraser~\cite{diffueraser}, it suffers from twice the inference time and poorer background reconstruction in Fig.~\ref{fig:ablation3}. Consequently, we leverage a mix of both models to generate our Remove and Add data, balancing their respective strengths and weaknesses.

\noindent \textbf{Ablations on Human Alignment with VLMs.} A total of 300 pairs were stratified across all 8 edit categories and cover outputs from multiple editing models and quality levels. Each pair was independently scored by \textit{3 annotators} using the same three 1--5 criteria. The final human score is the average of the three annotations. The inter-rater agreement is strong, with a two-way random-effects {ICC(2,k)=0.948} over the annotators' overall scores.
We compute PLCC and SROCC between the VLM overall scores and the mean human overall scores in Tab.~\ref{tab:vlmabl}. Seed1.6-VL shows the strongest agreement with humans ({PLCC/SROCC: 0.833/0.842}), while Gemini2.5-Pro also clearly outperforms InternVL3.5 and Qwen3-VL. This supports our benchmark design: \textit{stronger VLMs are used for final model ranking}, while \textit{open-source VLMs are used only as auxiliary references or coarse filters}. These results indicate that OpenVE-Bench is reasonably aligned with human judgment.

\begin{table}[t]
\centering
\renewcommand{\arraystretch}{1}
\setlength\tabcolsep{1pt}
\caption{Human--VLM alignment on OpenVE-Bench.}
\resizebox{0.75\linewidth}{!}{
\begin{tabular}{c|cccc}
\toprule
Judge & Seed1.6-VL & Gemini2.5-Pro & Intern3.5-VL-38B & Qwen3-VL-30B-A3B \\
\midrule
PLCC & \textbf{0.833} & 0.731 & 0.670 & 0.583 \\
SROCC & \textbf{0.842} & 0.756 & 0.643 & 0.555 \\
\bottomrule 
\end{tabular}}
\label{tab:vlmabl}%
\end{table}

\FloatBarrier

%% file: sec/7_conclusion.tex
\section{Conclusion} \label{sec:conclusion}
We introduce OpenVE-3M, a large-scale, high-quality dataset for instruction-following video editing. It comprises two main categories, spatially-aligned and non-spatially-aligned, encompassing a total of eight sub-categories. The detailed data construction and filtering pipelines are open-sourced to facilitate further research in the community. We also propose the OpenVE-Edit model, trained on OpenVE-3M. With only 5B parameters, our model outperforms a 14B parameter baseline. For standardized evaluation, we introduce OpenVE-Bench, which demonstrates high correlation with human judgment. On this benchmark, our method achieves SOTA performance among open-source models.

\noindent \textbf{Limitations and Future Work.} 
While our OpenVE-3M dataset is diverse, it omits certain categories like reference-based style transfer. Also, computational constraints limited our experiments to smaller-scale model architectures.
We plan to explore architectures that better unify spatially-aligned and non-spatially-aligned edits, potentially by adapting computationally-demanding image editing paradigms like noise concatenation along token dimensions, which remains under-explored in video. We will also investigate unified models, focusing on the bidirectional synergy where model understanding, generation and editing capabilities mutually enhance each other.

\section*{Acknowledgements}
This work was supported by Jianbing Lingyan Foundation of Zhejiang Province, P.R. China (Grant No. 2023C01022) and Major Project of Science and Technology of Yunnan Province, China under Grant 202402AD080001.

%% file: sec/X_suppl.tex
\renewcommand\thefigure{\Alph{section}\arabic{figure}}
\renewcommand\thetable{\Alph{section}\arabic{table}}  
\renewcommand\theequation{\Alph{section}\arabic{equation}}
\setcounter{equation}{0}
\setcounter{table}{0}
\setcounter{figure}{0}
\appendix

\section{Overview} 
\label{sec:overview}
The supplementary material presents the following sections to strengthen the main manuscript:

\begin{itemize}
\item[—] \textbf{Sec.}~\ref{sec:related_work} presents a detailed review of Related Works.
\item[—] \textbf{Sec.}~\ref{sec:sup_ppl} details the Data Construction Process with Workflows.
\item[—] \textbf{Sec.}~\ref{sec:sup_dataset} provides additional Statistics and Visualizations of the OpenVE-3M dataset.
\item[—] \textbf{Sec.}~\ref{sec:sup_model} shows further Quantitative and Qualitative Comparisons with SoTA methods.
\item[—] \textbf{Sec.}~\ref{sec:sup_bench} describes the detailed Prompt Design for OpenVE-Bench.
\end{itemize}

\section{Related Work}
\label{sec:related_work}

\subsection{Instruction-guided Image and Video Editing Datasets}
The domain of instruction-guided image editing has seen a proliferation of datasets in recent years. Early efforts, such as InstructPix2Pix\cite{instructpix2pix}, MagicBrush\cite{magicbrush}, HQ-Edit\cite{hqedit}, and AURORA\cite{aurora}, were limited in scale (tens to hundreds of thousands of samples) and quality, constrained by the generative models of their time. Subsequent research introduced more sophisticated data generation pipelines, leading to larger-scale and higher-fidelity datasets like SEED-Data-Edit\cite{seeddataedit}, OmniEdit\cite{omniedit}, AnyEdit\cite{anyedit}, UltraEdit\cite{ultraedit}, and SuperEdit\cite{superedit}. More recently, the advent of powerful foundational models, both open-source and proprietary, has further elevated the quality of editing data. Datasets including ImgEdit\cite{imgedit}, ShareGPT-4o-Image\cite{sharegpt4oimage}, GPT-Image-Edit\cite{gpteditimage}, and X2Edit\cite{x2edit} leverage these advanced models to produce state-of-the-art editing pairs.

However, the landscape of Instruction-guided video editing datasets is comparatively underdeveloped. InsV2V\cite{insv2v} extended Prompt-to-Prompt to video but inherited its quality limitations. VIVID-10M\cite{vivid} offers a large scale of 10 million samples but is restricted to three edit types and only provides masks instead of final edited videos, hindering its use for training end-to-end models. InsViE-1M\cite{insvie} generates video pairs by editing the initial frame and leveraging an image-to-video model, but struggles to maintain motion consistency with the source video. Señorita-2M\cite{senorita} trains multiple diffusion expert models for four semantic edit categories, yet its quality is capped by the performance of these specialized models. In summary, a significant gap persists in the availability of high-quality, large-scale video editing datasets comparable to those in the image domain.

\subsection{Instruction-guided Image and Video Editing Methods}
Recent advances in diffusion-based generative models, including text-to-image, text-to-video, and image-to-video systems, have catalyzed extensive research in editing. State-of-the-art instruction-guided image editing models are predominantly data-driven, fine-tuning pre-trained T2I models on large-scale editing datasets. A common architectural motif involves concatenating encoded image features (from VAE or SigLIP\cite{siglip}) with noise features. This concatenation is performed either along the sequence dimension, as seen in ImgEdit\cite{imgedit}, Step1X-Edit\cite{step1xedit}, FLUX-Kontext\cite{kontext}, OmniGen2\cite{omnigen2}, and Qwen-Image-Edit\cite{qwenimage}, or along the channel dimension, as in X2Edit\cite{x2edit}.

Following this trend, instruction-guided video editing models have recently proliferated. Omni-Video\cite{omnivideo} injects concatenated video and text features into a DiT network via cross-attention. Lucy-Edit\cite{lucyedit} concatenates VAE-encoded video features with noise along the channel dimension. InstructX\cite{instructx} employs LoRA to fine-tune an MLLM and adds noise to video features. UniVideo\cite{univideo} and ICVE\cite{icve} both adopt an MMDiT-like architecture, concatenating VAE-encoded video features with noise along the sequence dimension. Despite these varied implementations, a model that can effectively and efficiently handle multi-task, complex instructions for video editing has yet to emerge.

\subsection{Instruction-guided Image and Video Editing Benchmarks}
The escalating capabilities of editing models necessitate more sophisticated evaluation methodologies. Early research relied on generic similarity scores such as CLIP score\cite{clip}, PSNR\cite{psnr}, and SSIM\cite{ssim}. However, these metrics often fail to directly capture the perceptual quality and semantic correctness of an edit. More recently, studies\cite{imgedit,step1xedit,complextedit} have leveraged powerful VLMs like GPT-4o for nuanced image editing evaluation, scoring edits across multiple dimensions based on system prompts.

The evaluation of video editing is still in an exploratory phase. UNIC\cite{unic} uses a combination of metrics, including CLIP\cite{clip}/DINO\cite{dino} scores and VBench's\cite{vbench} quality indicators. VIE-Bench\cite{instructx} employs GPT-4o to assess sampled frames and overall video quality. EditVerse\cite{editverse} proposes a four-faceted evaluation including VLM assessment and text-video alignment. While IVEBench~\cite{icve} evaluates models on Video Quality, Instruction Compliance, and Video Fidelity, its evaluation process is time-consuming and places excessive emphasis on video quality over instruction adherence. Nevertheless, a universally accepted and effective evaluation protocol for video editing remains an open challenge.

\section{Detailed Data Piplines}
\label{sec:sup_ppl}
\setcounter{figure}{0}
\setcounter{table}{0}
\setcounter{equation}{0}
\begin{figure*}[t!]
\centering
\includegraphics[width=1\linewidth]{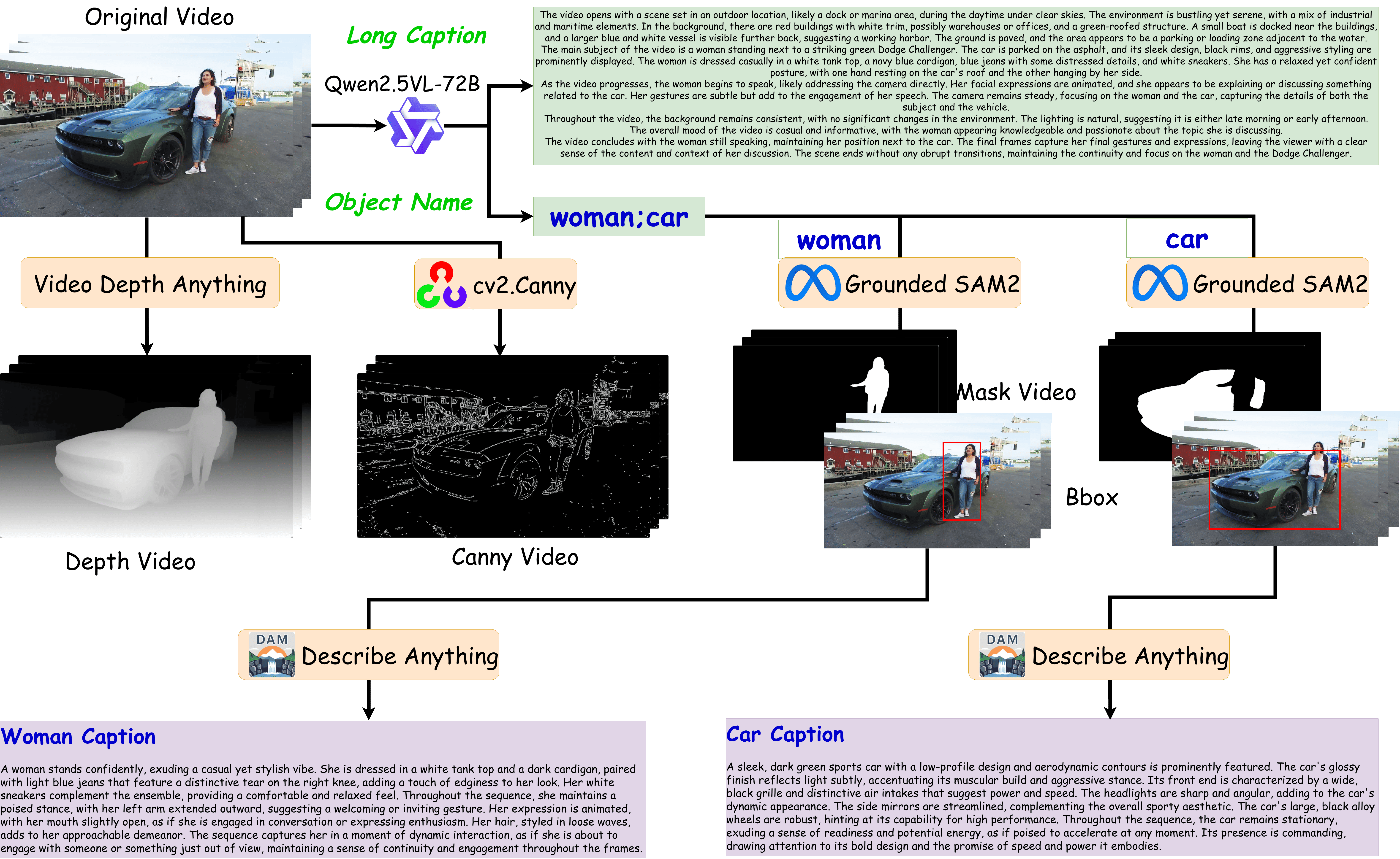}
\caption{A detailed flowchart example of the Stage 1 data pipeline.}
\label{fig:sup_stage1_ppl1}
\end{figure*}
Fig.~\ref{fig:sup_stage1_ppl1} illustrates the detailed data preprocessing workflow for Stage 1. The process begins with a video clip, for instance, frames 65 to 129, extracted from an open-source video dataset~\cite{opensoraplan,openvid,ultravideo} and resized to 720P resolution. This clip, along with a meticulously crafted prompt, is fed into the Qwen2.5VL-72B~\cite{qwen25vl} model to generate a comprehensive video description and a list of object names.
Concurrently, the Video Depth Anything~\cite{videodepthanything} model and the OpenCV~\cite{opencv} Canny algorithm are employed to produce the corresponding depth and Canny videos, respectively. With the object names identified, Grounded SAM2~\cite{groundedsam2} is then applied to each object to perform detection and segmentation, yielding its bounding box and mask video.
Finally, the bounding box and mask video for an object serve as input to the Describe Anything~\cite{describeanything} model, which generates a detailed local description (e.g., "Woman Caption" and "Car Caption" as shown in the figure). This entire pipeline is applied to all videos, and all processed data types are saved to facilitate straightforward retrieval during the construction of Stage 2.

\begin{figure*}[t!]
\centering
\includegraphics[width=0.82\linewidth]{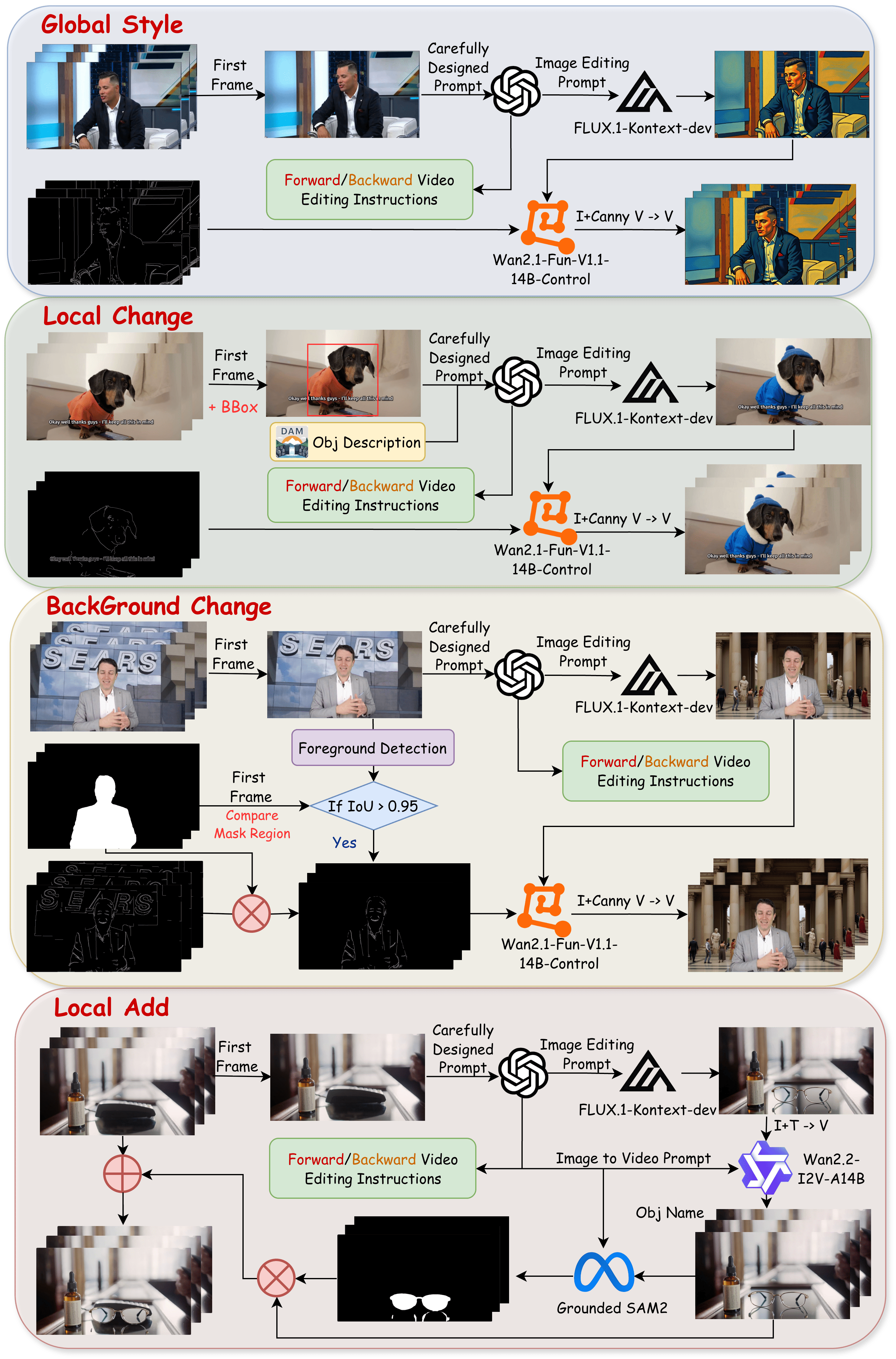}
\caption{Detailed workflow of the four video editing data construction pipelines: Global Style, Local Change, Background Change, and Local Add.}
\label{fig:sup_ppl1}
\end{figure*}

\begin{figure*}[t!]
\centering
\includegraphics[width=0.85\linewidth]{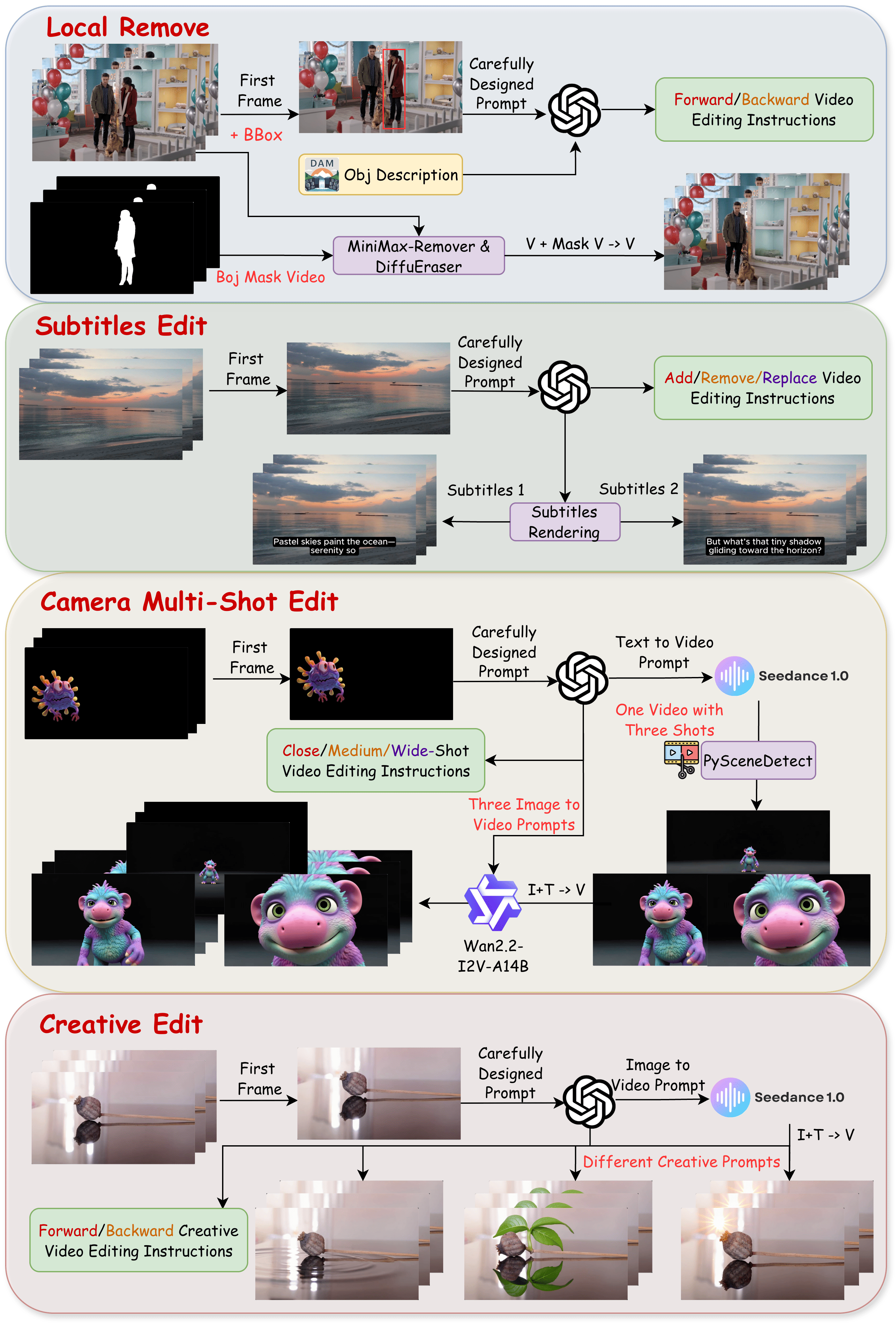}
\caption{Detailed workflow of the four video editing data construction pipelines: Local Remove, Subtitles Edit, Camera Multi-shot Edit, and Creative Edit.}
\label{fig:sup_ppl2}
\end{figure*}

We present the flowchart of the data construction pipeline for instruction-following video editing. Fig.~\ref{fig:sup_ppl1} illustrates the construction process for four categories: Global Style, Local Change, Background Change, and Local Add. Fig.~\ref{fig:sup_ppl2} shows the process for Local Remove, Subtitles Edit, Camera Multi-shot Edit, and Creative Edit. Detailed descriptions of the construction process are provided in Sec.~\textcolor{blue}{2}.

\section{OpenVE-3M Dataset Statistics and Visualization}
\label{sec:sup_dataset}
\setcounter{figure}{0}
\setcounter{table}{0}
\setcounter{equation}{0}

\noindent\textbf{Detailed statistics of OpenVE-3M.}
This section presents the detailed statistics in the OpenVE-3M dataset, as shown in Tab.~\ref{tab:sup_data}.

\begin{table}[htbp]
    \centering
    \caption{The detailed statistics of the OpenVE-3M dataset.}
    \renewcommand{\arraystretch}{1.0}
    \setlength\tabcolsep{5.0pt}
    \resizebox{0.5\linewidth}{!}{
        \begin{tabular}{c|cc}
        \toprule[0.1em]
        Category & Video Paris & Ratio \\
    \midrule
      Global Style   &   431716  & 14.26\% \\
    Background Change  &   396212 & 13.09\%  \\
      Local Change   &  478696  & 15.85\% \\
      Local Add   &   400595 & 13.24\% \\
      Local Remove   &   337541  & 11.15\%\\
      Subtitles Edit   &   400000 & 13.21\% \\
    \midrule
      Camera  Edit   &   381045 & 12.59\% \\
      Creative Edit   &   200042 & 6.61\% \\
        \toprule[0.1em]
        \end{tabular}
    }
    \label{tab:sup_data}
\end{table}

\begin{figure*}[htpb]
\centering
\includegraphics[width=0.85\linewidth]{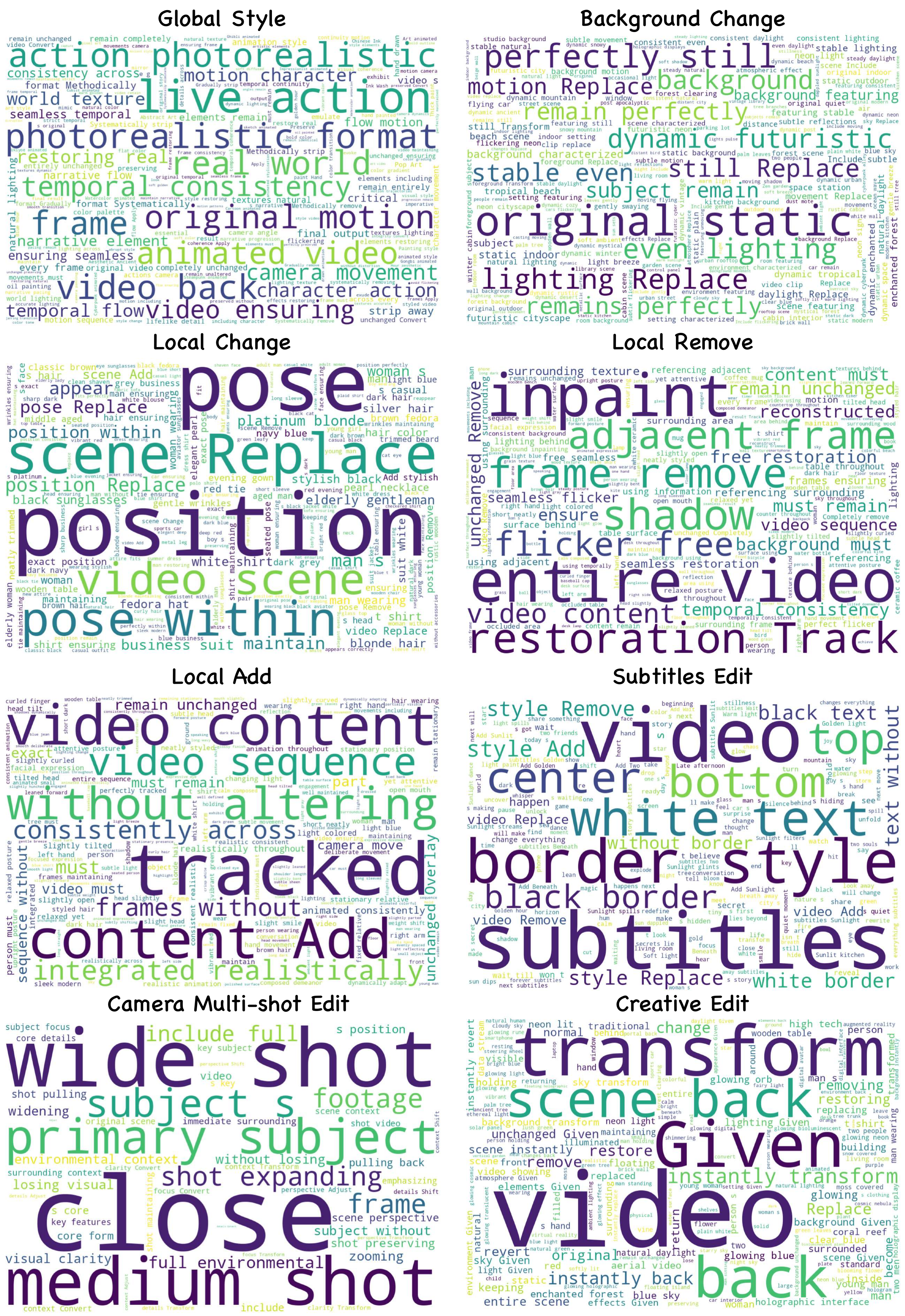}
\caption{Word clouds of different categories in OpenVE-3M Dataset.}
\label{fig:sup_word_cloud}
\end{figure*}

\noindent\textbf{Visualization of the Word Cloud for Different Video Editing Prompt.}
This section presents word clouds of the editing instructions for all categories in the OpenVE-3M dataset, as shown in Fig.~\ref{fig:sup_word_cloud}.

\begin{figure*}[htpb]
\centering
\includegraphics[width=0.79\linewidth]{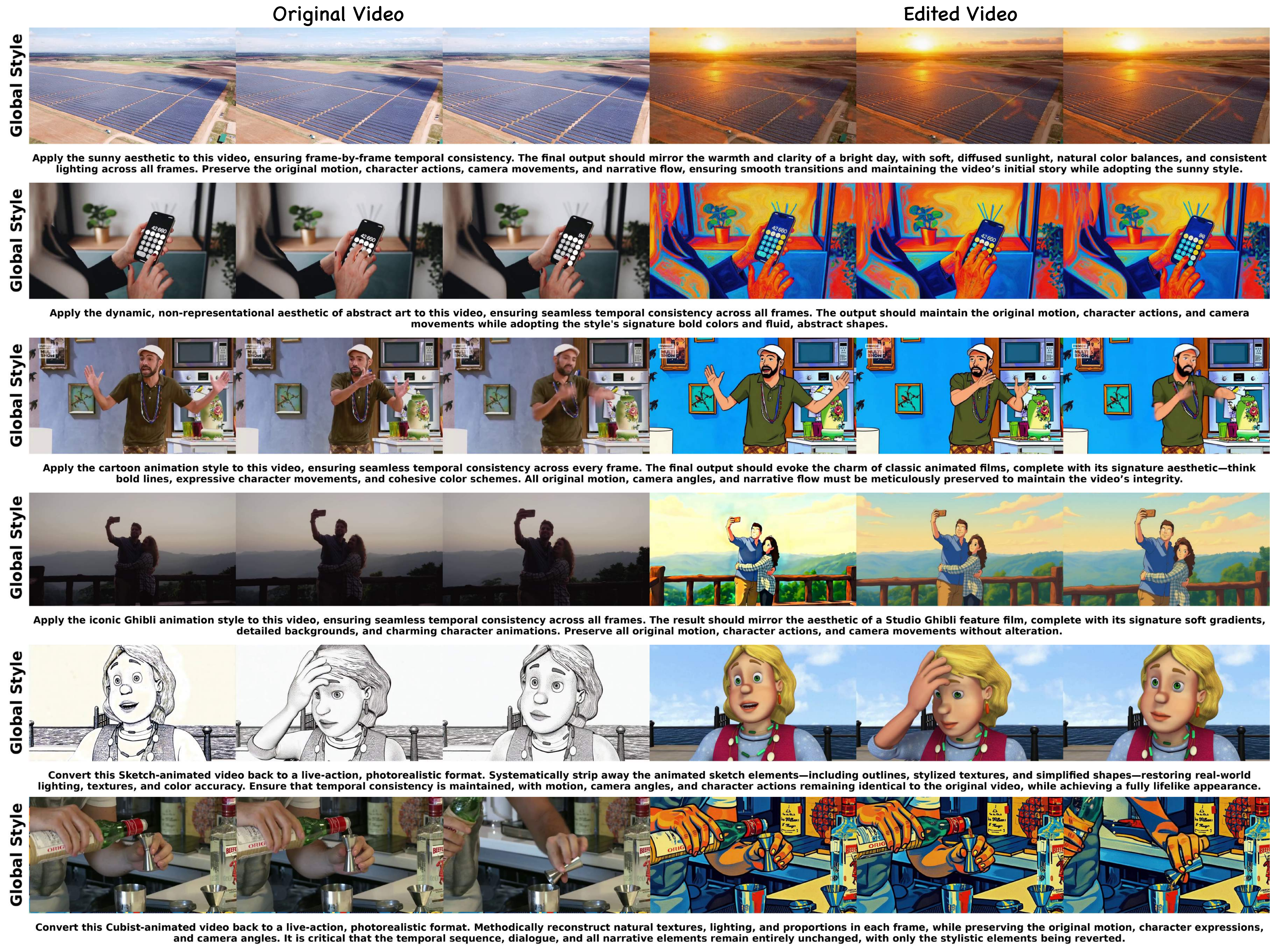}
\caption{Qualitative Visualization Results of \textbf{Global Style} category in OpenVE-3M Dataset.}
\label{fig:sup_data_gs}
\end{figure*}

\begin{figure*}[htpb]
\centering
\includegraphics[width=0.79\linewidth]{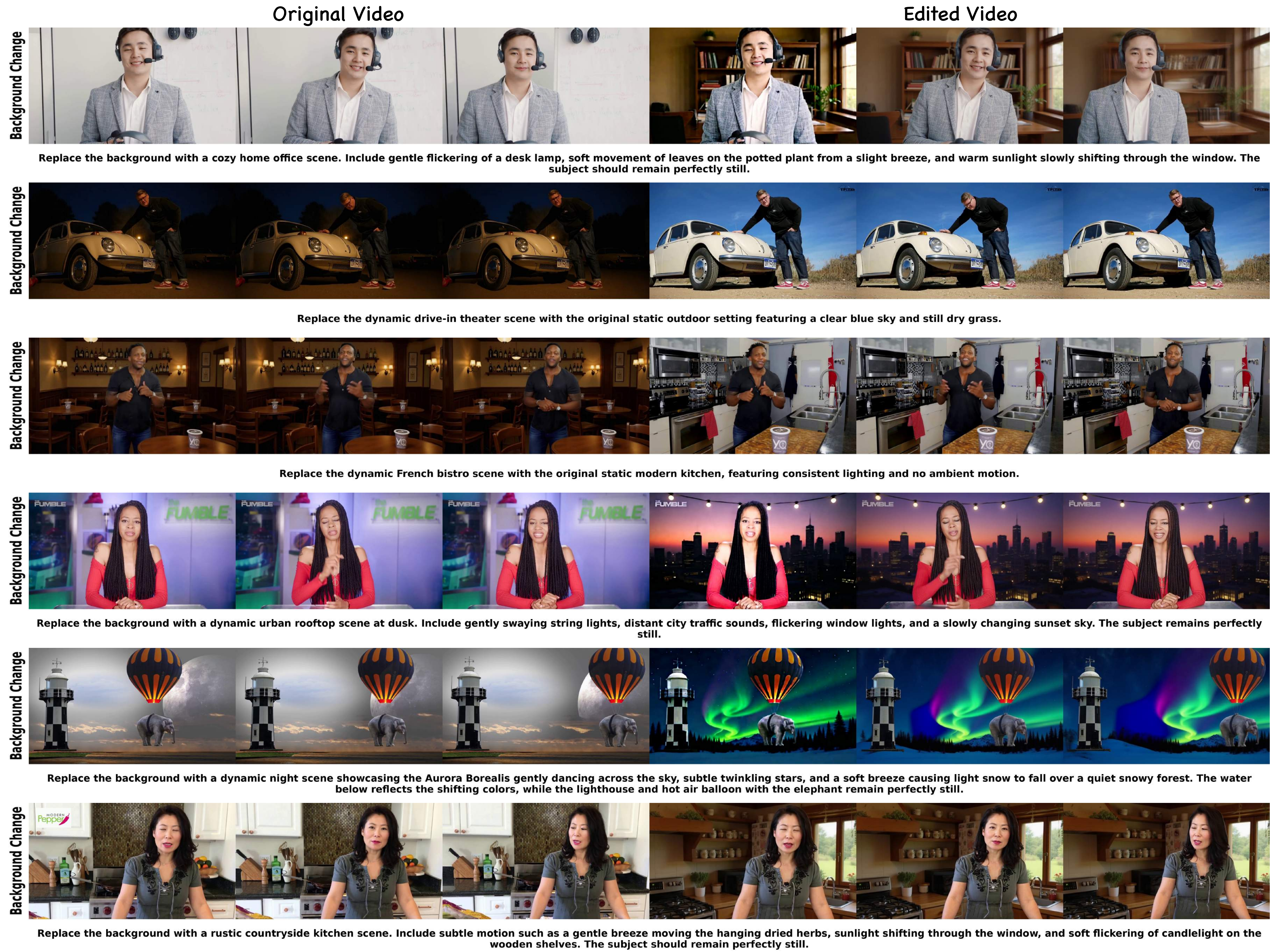}
\caption{Qualitative Visualization Results of \textbf{Background Change} category in OpenVE-3M Dataset.}
\label{fig:sup_data_bc}
\end{figure*}

\begin{figure*}[htpb]
\centering
\includegraphics[width=0.79\linewidth]{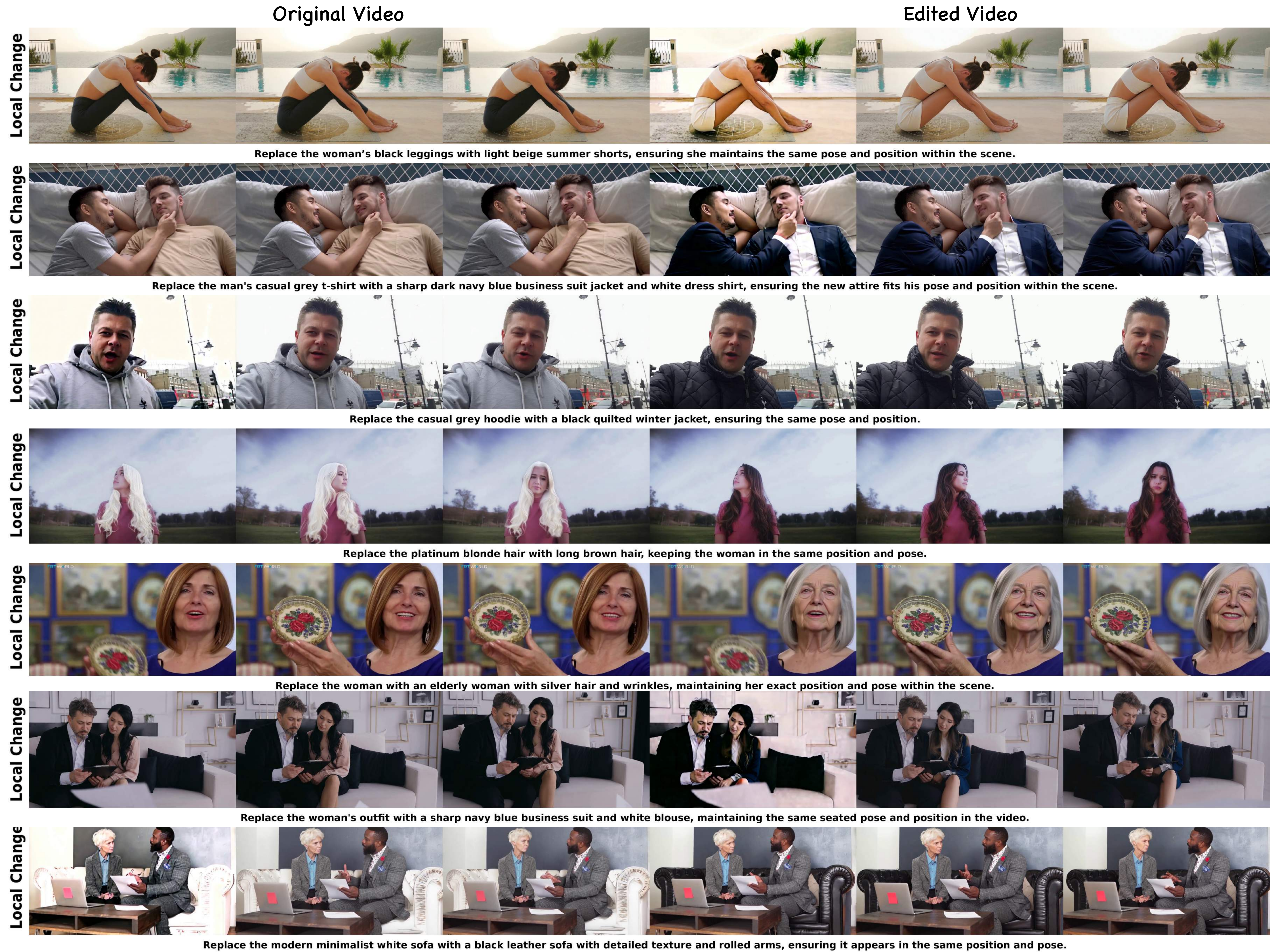}
\caption{Qualitative Visualization Results of \textbf{Local Change} category in OpenVE-3M Dataset.}
\label{fig:sup_data_lc}
\end{figure*}

\begin{figure*}[htpb]
\centering
\includegraphics[width=0.79\linewidth]{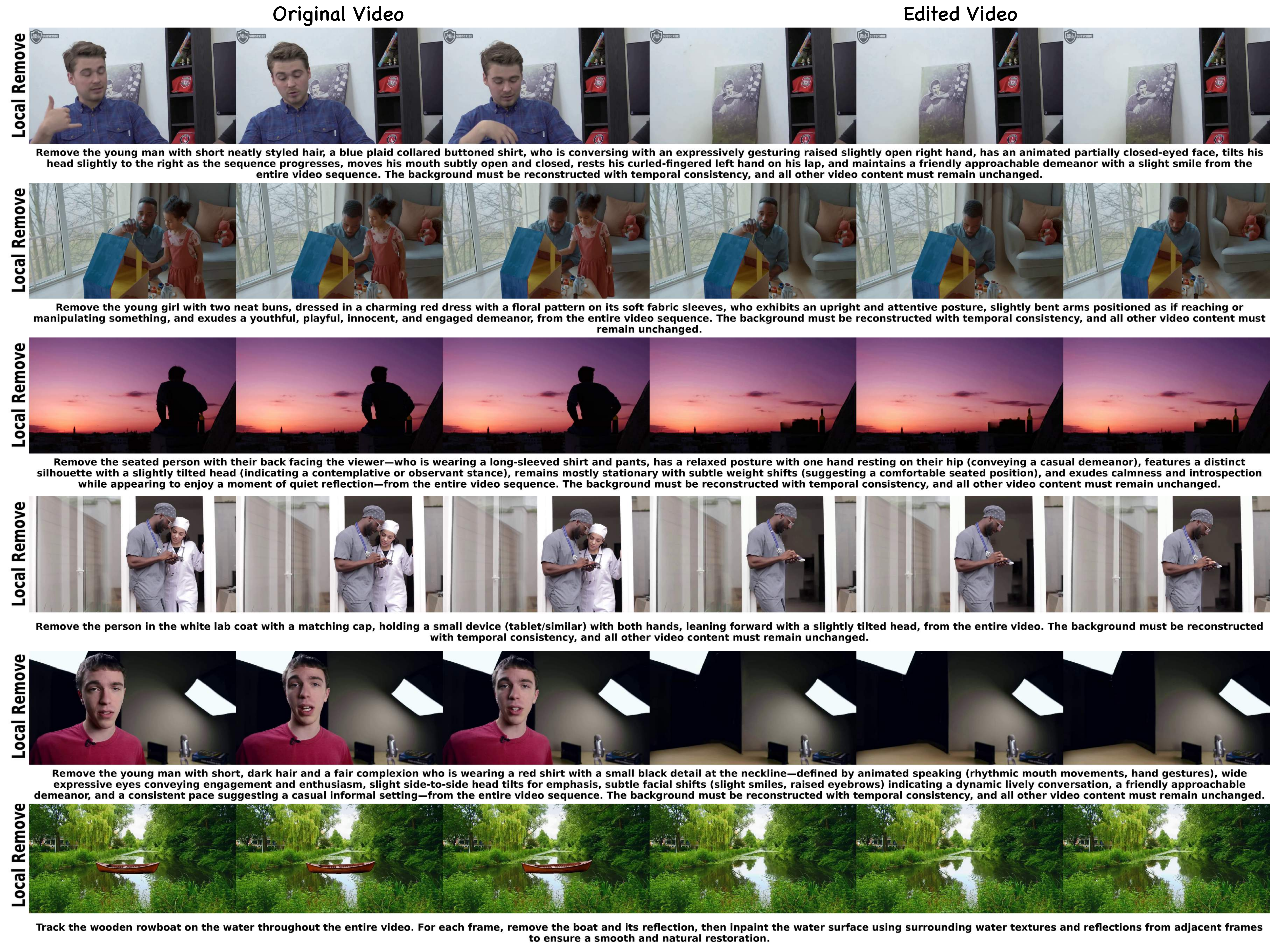}
\caption{Qualitative Visualization Results of \textbf{Local Remove} category in OpenVE-3M Dataset.}
\label{fig:sup_data_lr}
\end{figure*}

\begin{figure*}[htpb]
\centering
\includegraphics[width=0.79\linewidth]{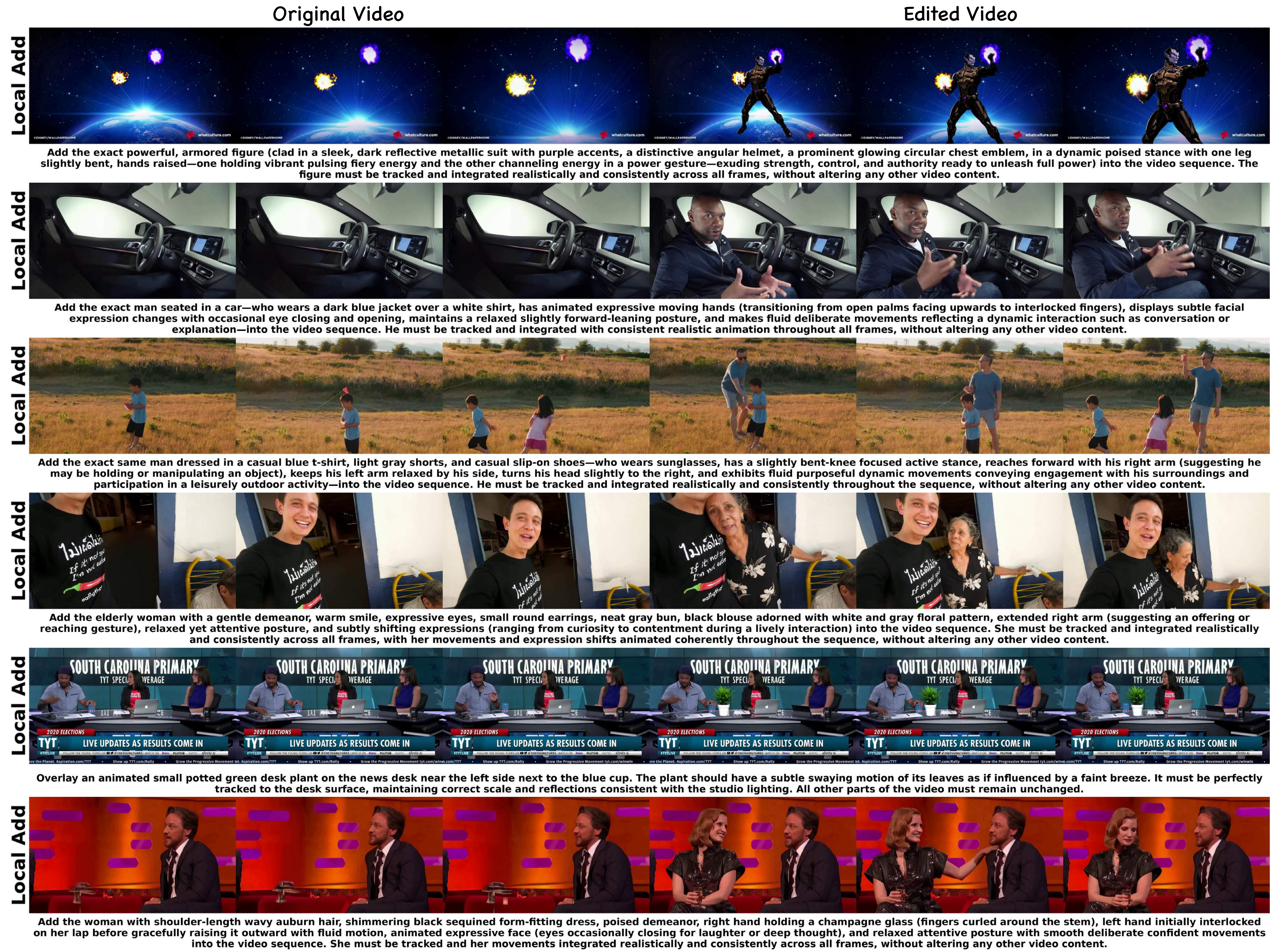}
\caption{Qualitative Visualization Results of \textbf{Local Add} category in OpenVE-3M Dataset.}
\label{fig:sup_data_la}
\end{figure*}

\begin{figure*}[htpb]
\centering
\includegraphics[width=0.79\linewidth]{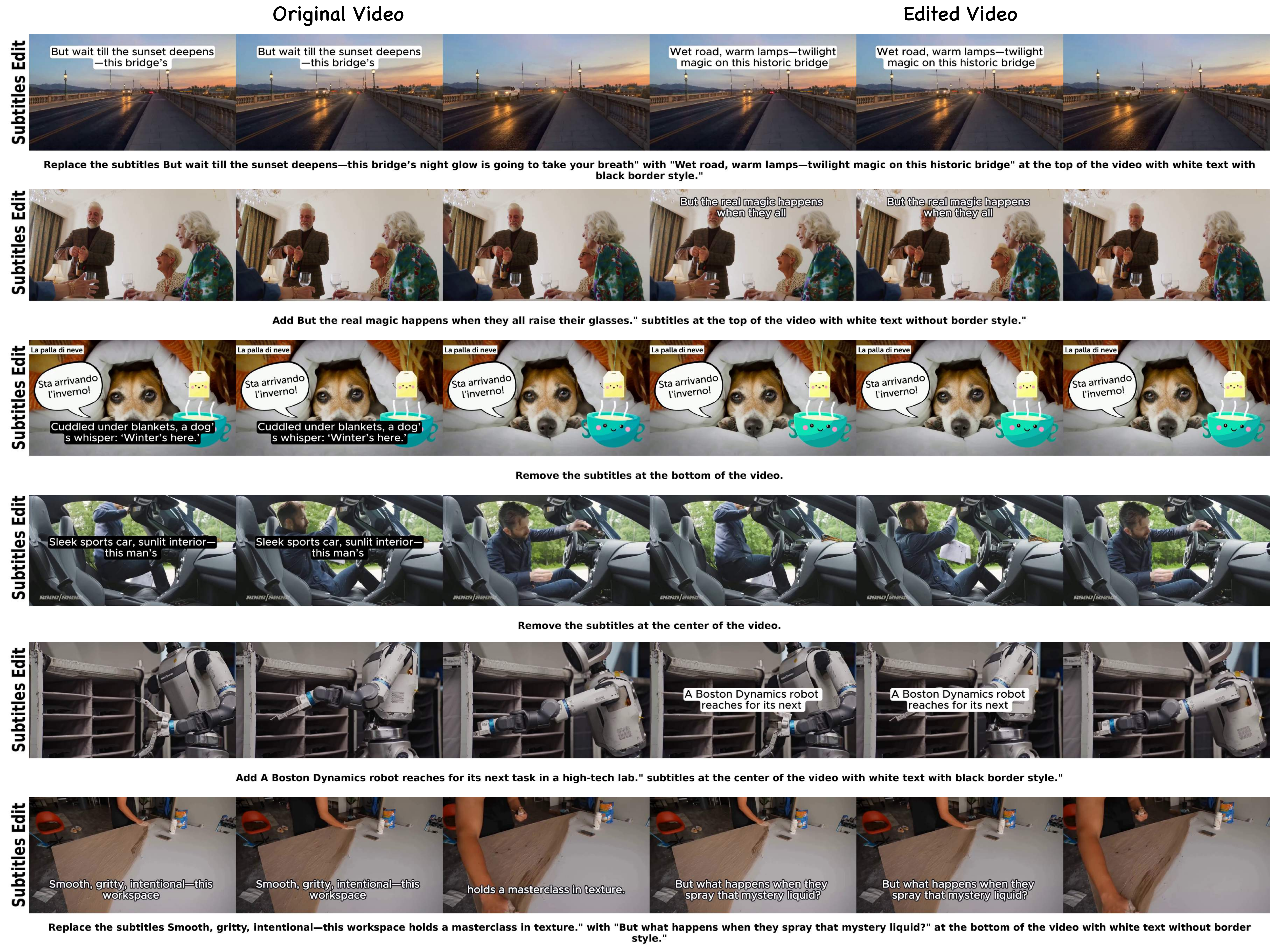}
\caption{Qualitative Visualization Results of \textbf{Subtitles Edit} category in OpenVE-3M Dataset.}
\label{fig:sup_data_se}
\end{figure*}

\begin{figure*}[htpb]
\centering
\includegraphics[width=0.79\linewidth]{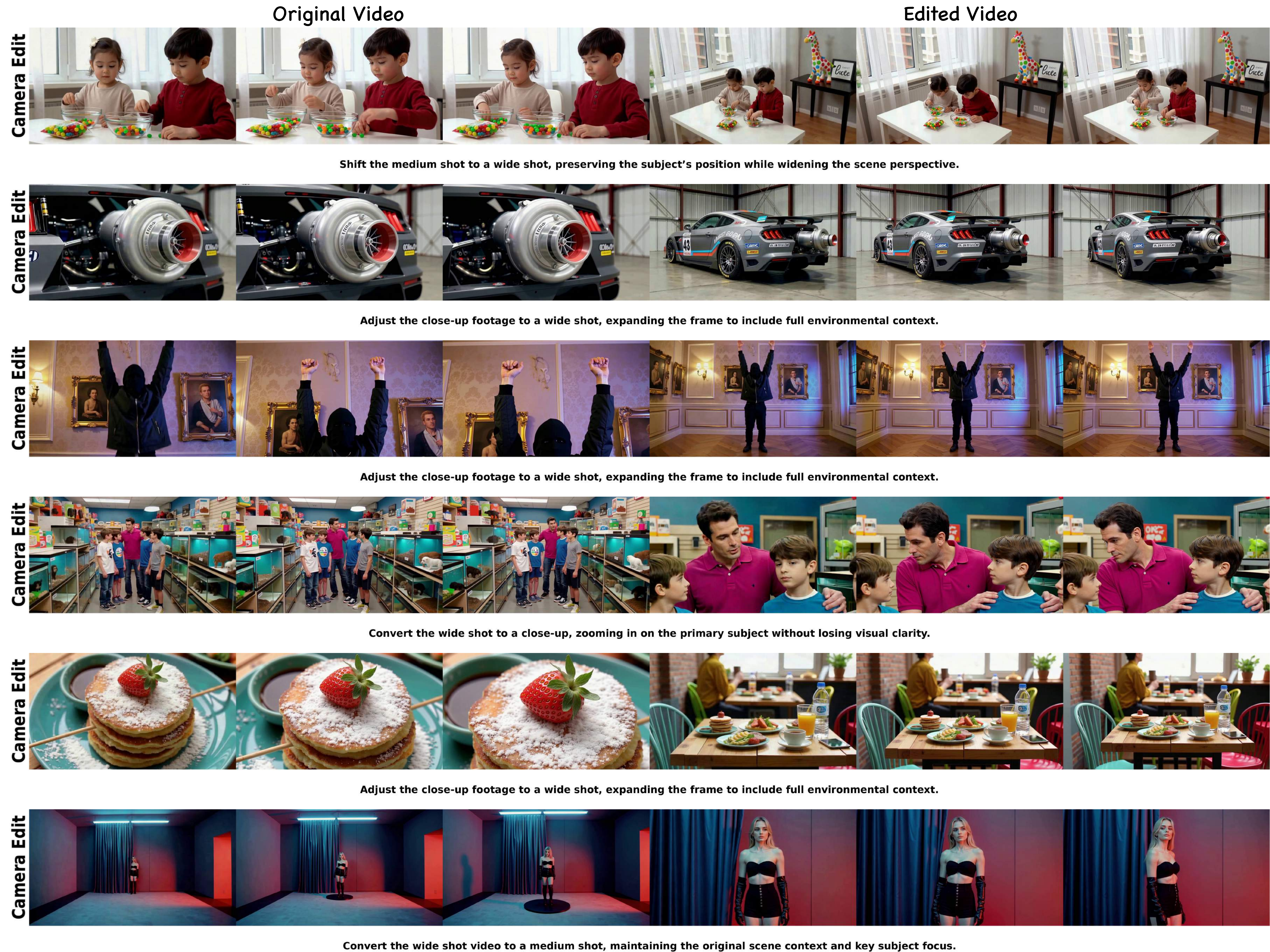}
\caption{Qualitative Visualization Results of \textbf{Camera Multi-Shot Edit} category in OpenVE-3M Dataset.}
\label{fig:sup_data_cme}
\end{figure*}

\begin{figure*}[htpb]
\centering
\includegraphics[width=0.79\linewidth]{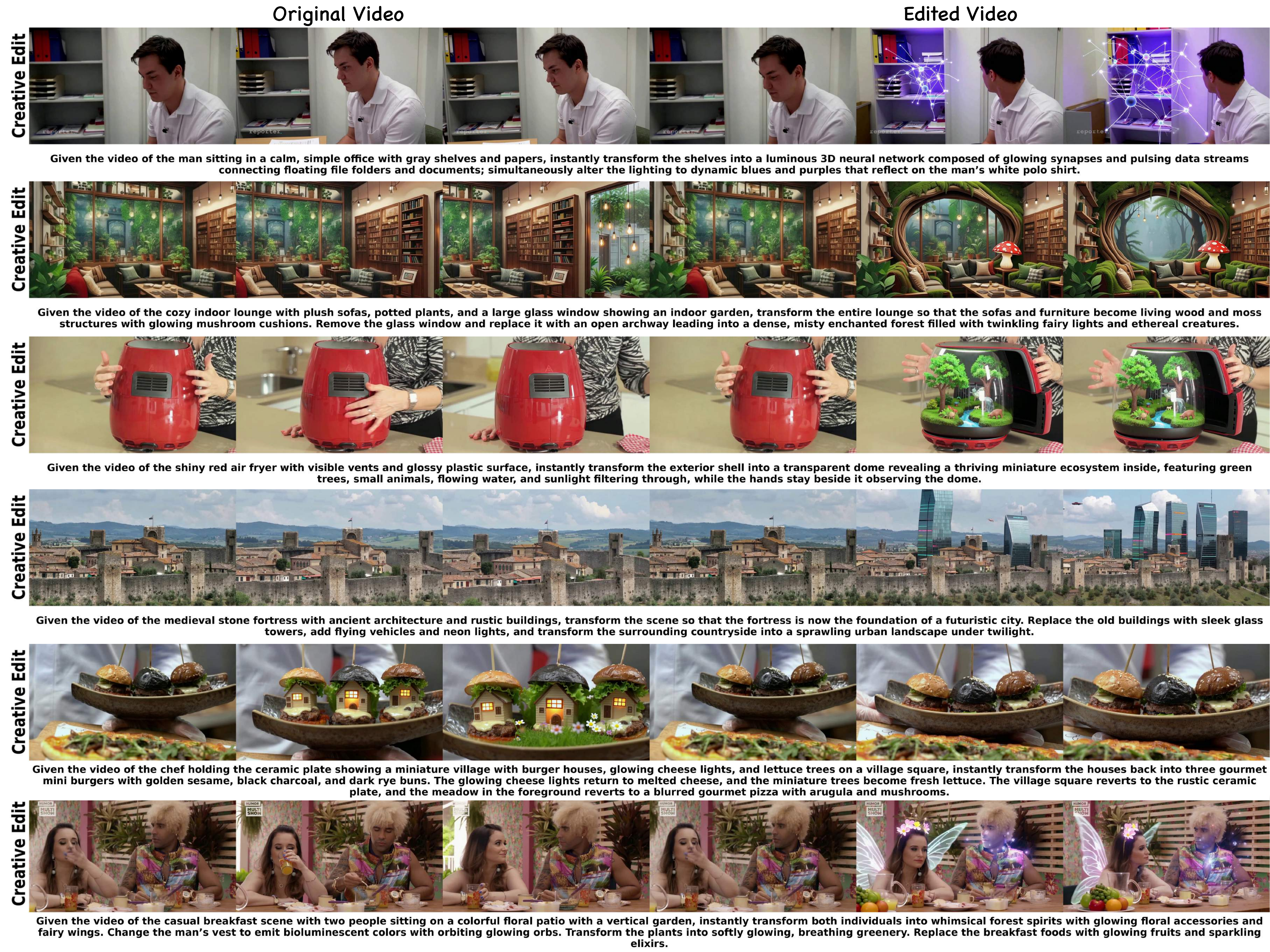}
\caption{Qualitative Visualization Results of \textbf{Creative Edit} category in OpenVE-3M Dataset.}
\label{fig:sup_data_cre}
\end{figure*}

\noindent\textbf{Visualization of Different Video Editing Categories.}
This section presents additional visual results for all categories in the OpenVE-3M dataset. Specifically, results for Global Style, Local Change, Background Change, Local Add, Local Remove, Subtitles Edit, Camera Multi-shot Edit, and Creative Edit are shown in Fig.~\ref{fig:sup_data_gs}, Fig.~\ref{fig:sup_data_bc}, Fig.~\ref{fig:sup_data_lc}, Fig.~\ref{fig:sup_data_lr}, Fig.~\ref{fig:sup_data_la}, Fig.~\ref{fig:sup_data_se}, Fig.~\ref{fig:sup_data_cme} and Fig.~\ref{fig:sup_data_cre} respectively.

\section{More Qualitative and Quantitative Comparison with SoTAs}
\label{sec:sup_model}
\setcounter{figure}{0}
\setcounter{table}{0}
\setcounter{equation}{0}
\noindent\textbf{Quantitative Comparison.}
Tab.~\ref{tab:sup_internvl}and \ref{tab:sup_qwen3vl} detail the quantitative results of our method compared to SoTA approaches on the InternVL3.5-38B~\cite{internvl35} and Qwen3-VL-32B~\cite{qwen25vl} backbones. Our method achieves superior performance among all evaluated open-source models on both platforms.
Notably, the average scores produced by InternVL3.5-38B~\cite{internvl35} and Qwen3-VL-32B~\cite{qwen25vl} are substantially higher than those from more capable models like Seed-1.6VL~\cite{seed15vl} and Gemini-2.5Pro~\cite{gemini25}. We attribute this score inflation to evaluation inaccuracies stemming from the inherent limitations of these two models. Two primary factors contribute to this phenomenon. First, their deficient video understanding capabilities lead to overly generous scoring. Second, our evaluation prompts include a critical constraint: scores for "Consistency \& Detail Fidelity" and "Visual Quality \& Stability" are not permitted to exceed the score for "Instruction Following". However, InternVL3.5-38B~\cite{internvl35} and Qwen3-VL-32B~\cite{qwen25vl} exhibit context-length limitations, such as instruction forgetting or misinterpretation. Consequently, they fail to adhere to this constraint, resulting in artificially elevated average scores.
Therefore, although the relative performance hierarchy remains discernible, we strongly advise using the more robust Seed-1.6VL~\cite{seed15vl} and Gemini-2.5-Pro~\cite{gemini25} models to obtain more accurate and reliable evaluations.

\begin{figure*}[htpb]
\centering
\includegraphics[width=1\linewidth]{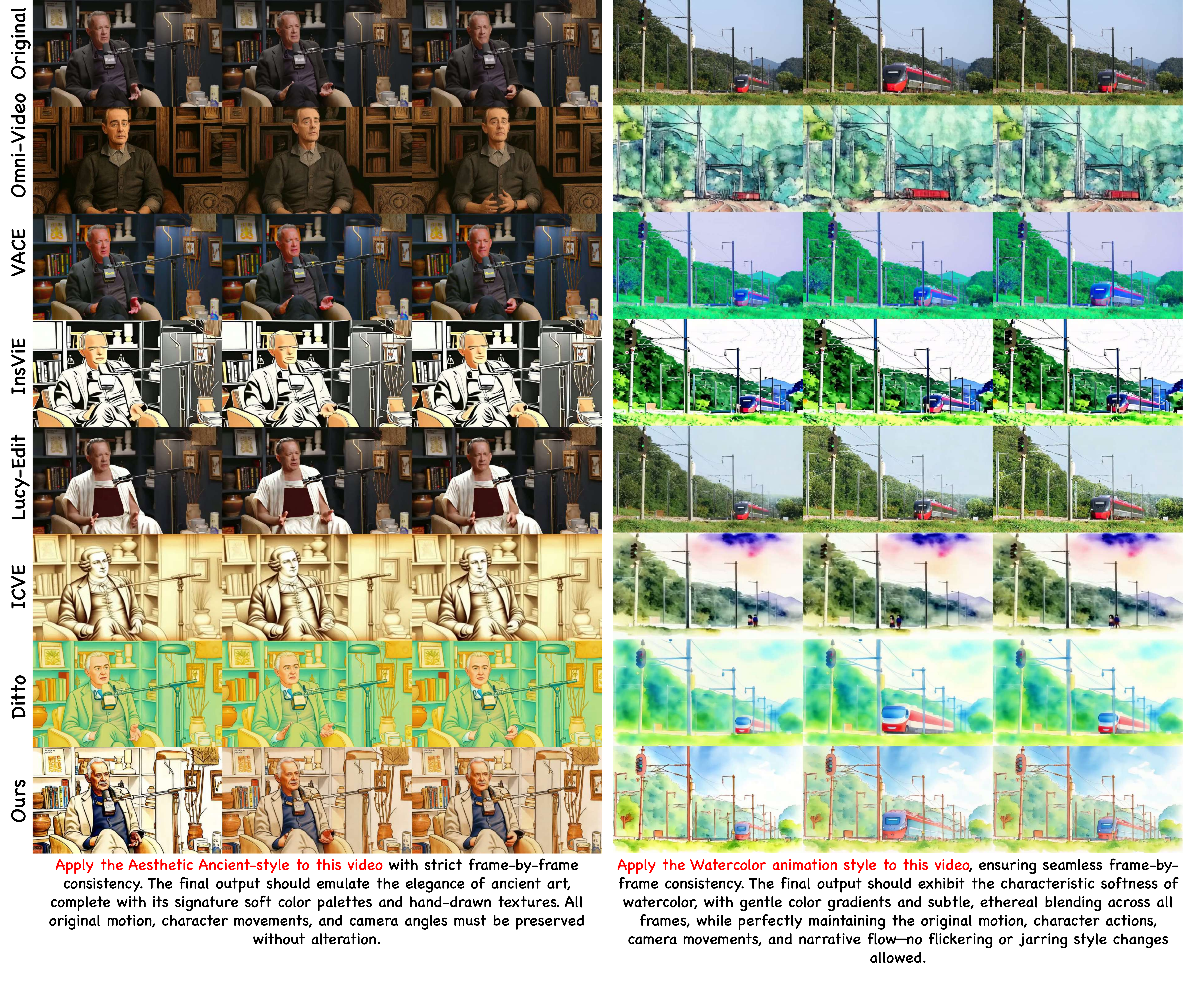}
\caption{Comparison of visual qualitative results with SoTAs in the \textbf{Global Style} category.}
\label{fig:sup_model_gs}
\end{figure*}

\begin{figure*}[htpb]
\centering
\includegraphics[width=1\linewidth]{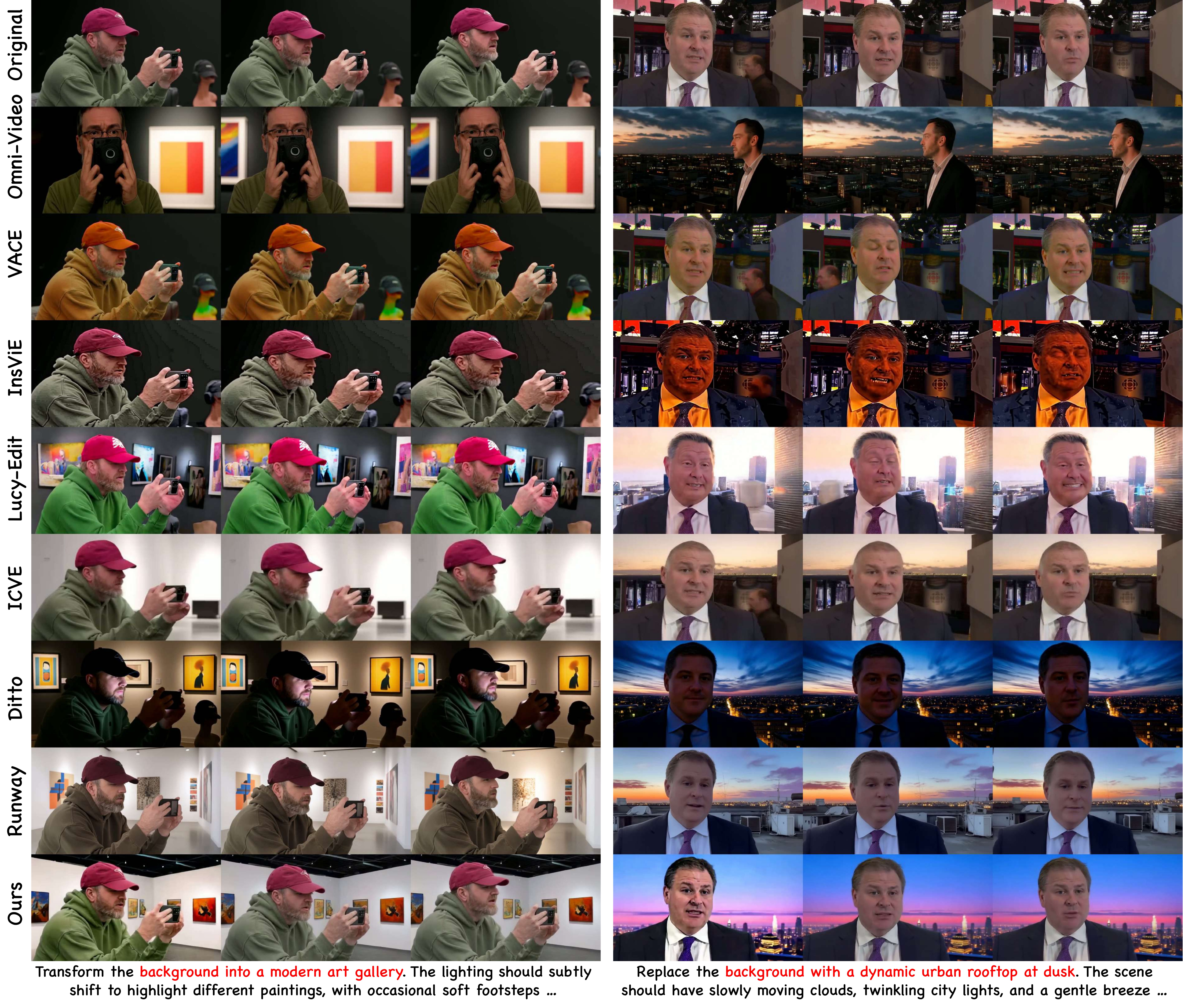}
\caption{Comparison of visual qualitative results with SoTAs in the \textbf{Background Change} category.}
\label{fig:sup_model_bc}
\end{figure*}

\begin{figure*}[htpb]
\centering
\includegraphics[width=1\linewidth]{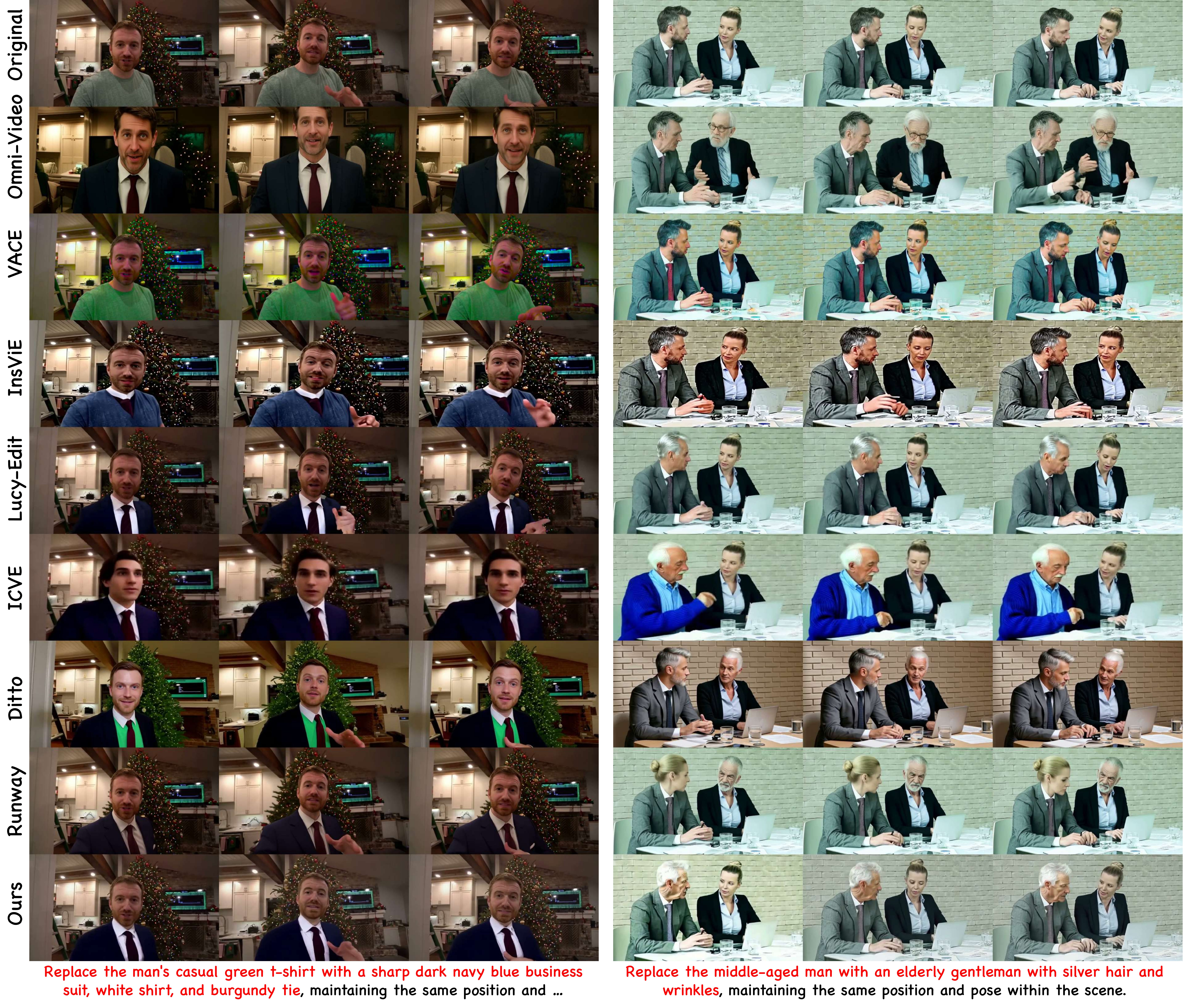}
\caption{Comparison of visual qualitative results with SoTAs in the \textbf{Local Change} category.}
\label{fig:sup_model_lc}
\end{figure*}

\begin{figure*}[htpb]
\centering
\includegraphics[width=1\linewidth]{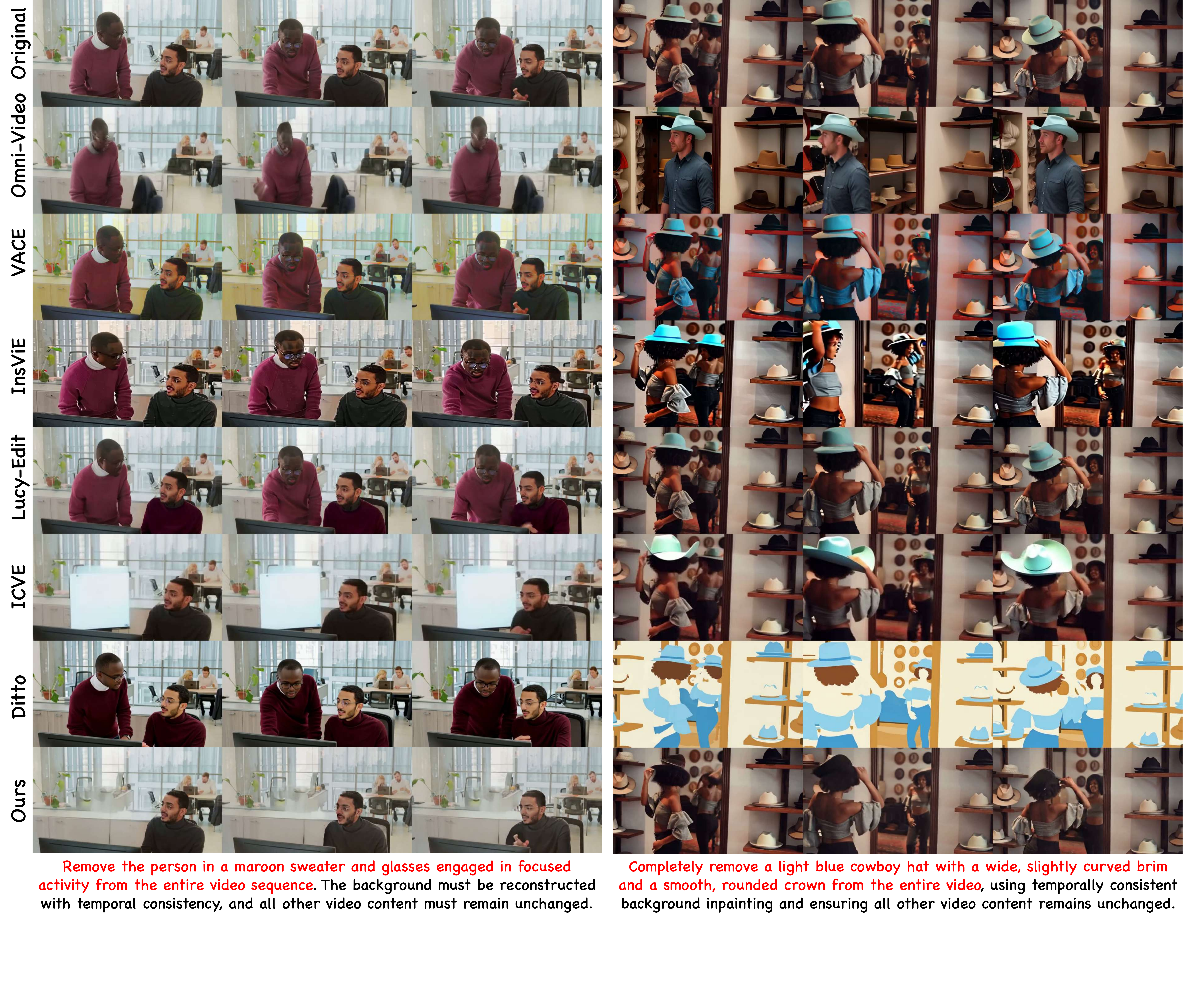}
\caption{Comparison of visual qualitative results with SoTAs in the \textbf{Local Remove} category.}
\label{fig:sup_model_lr}
\end{figure*}

\begin{figure*}[htpb]
\centering
\includegraphics[width=1\linewidth]{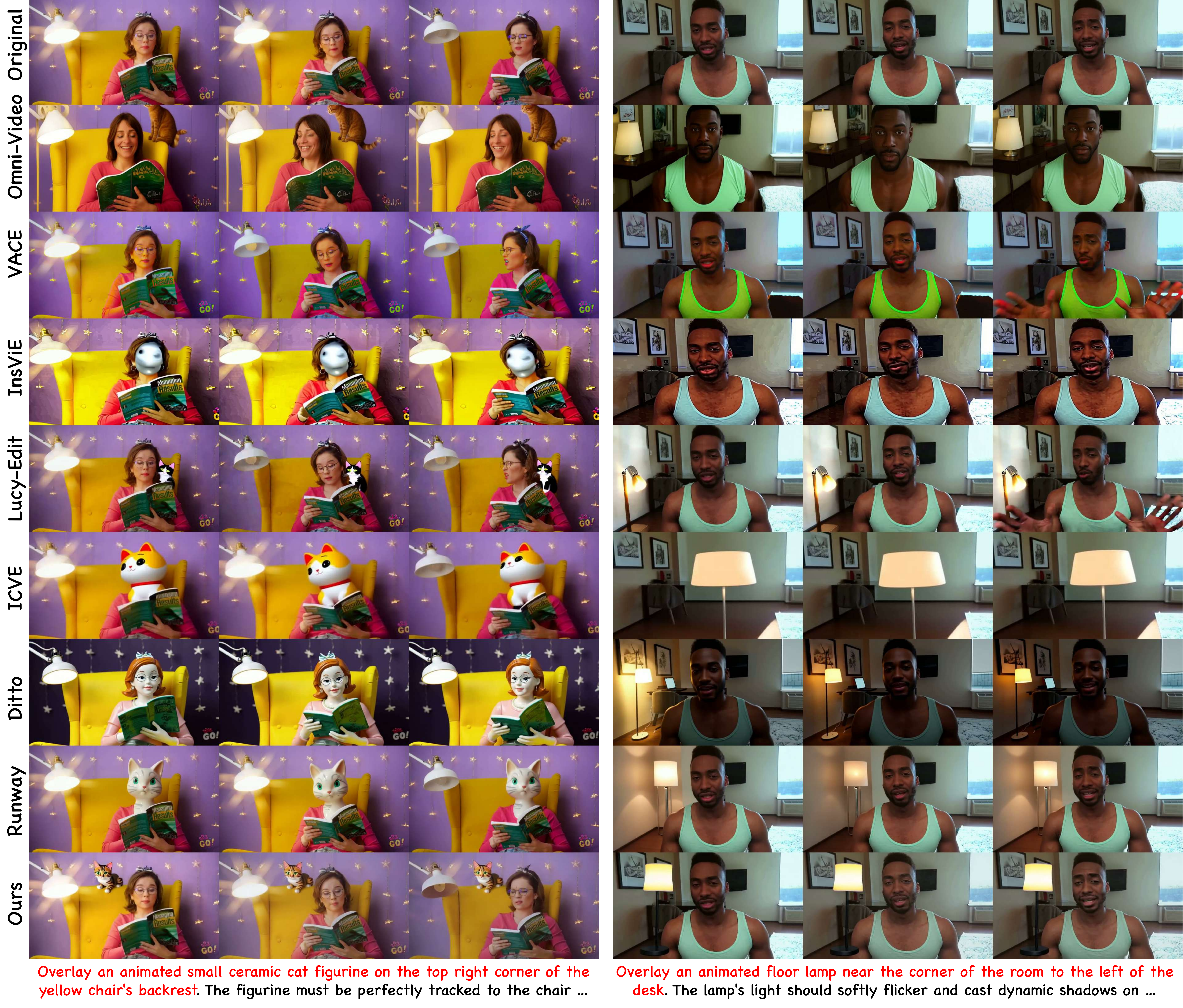}
\caption{Comparison of visual qualitative results with SoTAs in the \textbf{Local Add} category.}
\label{fig:sup_model_la}
\end{figure*}

\begin{figure*}[htpb]
\centering
\includegraphics[width=1\linewidth]{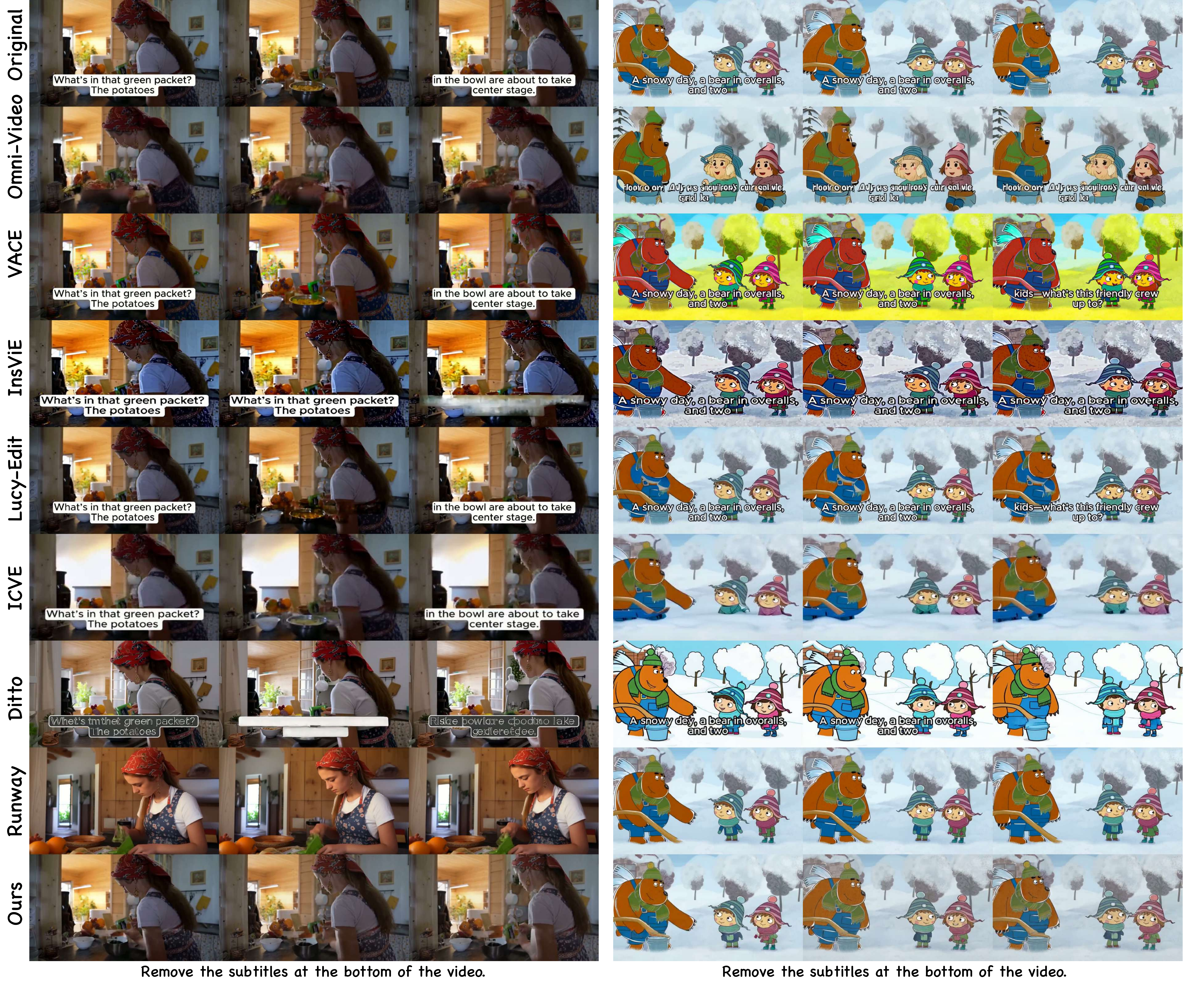}
\caption{Comparison of visual qualitative results with SoTAs in the \textbf{Subtitles Edit} category.}
\label{fig:sup_model_se}
\end{figure*}

\begin{figure*}[htpb]
\centering
\includegraphics[width=1\linewidth]{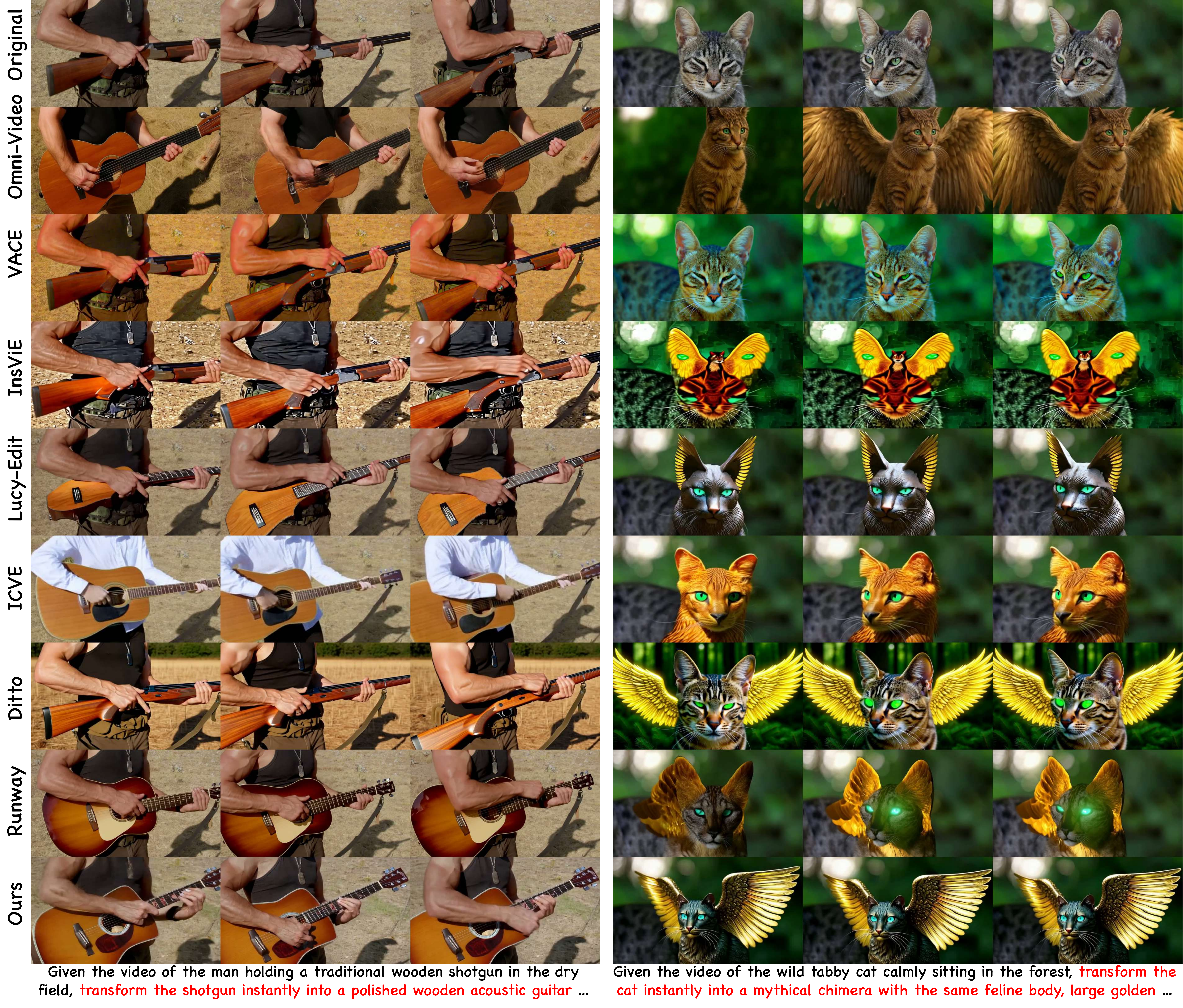}
\caption{Comparison of visual qualitative results with SoTAs in the \textbf{Creative Edit} category.}
\label{fig:sup_model_cre}
\end{figure*}

\noindent\textbf{Qualitative Comparison.}
This section provides further qualitative comparisons between our OpenVE-Edit method and existing SoTA methods in Fig.~\ref{fig:sup_model_gs}, Fig.~\ref{fig:sup_model_bc}, Fig.~\ref{fig:sup_model_lc}, Fig.~\ref{fig:sup_model_lr}, Fig.~\ref{fig:sup_model_la}, Fig.~\ref{fig:sup_model_se}, Fig.~\ref{fig:sup_model_cre}.

\section{OpenVE-Bench}
\label{sec:sup_bench}

\noindent\textbf{Detailed statistics of OpenVE-Bench.}
This section presents the detailed statistics in the OpenVE-Bench benchmark, as shown in Tab.~\ref{tab:sup_bench}.

\begin{table}[htbp]
    \centering
    \caption{The detailed statistics of the OpenVE-Bench dataset.}
    \renewcommand{\arraystretch}{1.0}
    \setlength\tabcolsep{5.0pt}
    \resizebox{0.4\linewidth}{!}{
        \begin{tabular}{c|c}
        \toprule[0.1em]
        Category & Video Paris \\
    \midrule
      Global Style   &   58 \\
    Background Change  &   59   \\
      Local Change   &  65  \\
      Local Add   &   67 \\
      Local Remove   &   59 \\
      Subtitles Edit   &   50 \\
    \midrule
      Camera  Edit   &   43  \\
      Creative Edit   &   30 \\
        \toprule[0.1em]
        \end{tabular}
    }
    \label{tab:sup_bench}
\end{table}

\noindent\textbf{Detailed Prompt Design of OpenVE-Bench.}
This section presents the detailed system prompts used for evaluation on OpenVE-Bench across each category. We design a unique system prompt for each category to guide the evaluation process. This process involves scoring the editing instruction, the source video, and the edited video on a 1-to-5 scale across three key aspects: Instruction Compliance, Consistency \& Detail Fidelity, and Visual Quality \& Stability. Specific design details for each editing category are provided on Global Style (Prompt.~\ref{prompt:gs}), Background Change (Prompt.~\ref{prompt:bc}), Local Change (Prompt.~\ref{prompt:lc}), Local Remove (Prompt.~\ref{prompt:lr}), Local Add (Prompt.~\ref{prompt:la}), Subtitles Edit (Prompt.~\ref{prompt:se}), Camera Multi-shot Edit (Prompt.~\ref{prompt:cme}), and Creative Edit (Prompt.~\ref{prompt:cre}).

\begin{table*}[htbp]
    \centering
    \renewcommand{\arraystretch}{1.0}
    \caption{Quantitative Comparison on \textbf{OpenVE-Bench} with InternVL3.5~\cite{internvl35}. \protect\sethlcolor{whit_tab}\hl{White}, \protect\sethlcolor{oran_tab}\hl{yellow}, \protect\sethlcolor{gray_tab}\hl{gray}, and \protect\sethlcolor{blue_tab}\hl{blue} backgrounds indicate Closed-source, Open-source and Ours, respectively. \#Params. and \#Reso. refers to Parameters and Resolutions, respectively.}
    \resizebox{1\linewidth}{!}{
        \begin{tabular}{c|cc|c|cccccccc}
        \toprule[0.1em]
Methods     & \#Params. & \#Reso.  & Overall & Global Style & Background Change & Local Change & Local Remove & Local Add & Subtitle Edit & Creative Edit & Camera Edit \\
    \midrule
    \rowcolor{oran_tab}Runway Aleph      & -         & 1280$\times$720 & 4.13 & 4.21 & 3.55 & 4.65 & 4.33 & 3.33 & 4.18 & 4.04 & 4.80 \\
\rowcolor{gray_tab}VACE~\cite{vace}        & 14B       & 1280$\times$720 & 2.15 & 2.89 & 1.54 & 2.64 & 1.92 & 1.24 & 2.64 & 2.57 & 2.49 \\
\rowcolor{gray_tab}OmniVideo~\cite{omnivideo}   & 1.3B      & 640$\times$352  & 1.41 & 1.57 & 1.09 & 1.00 & 1.00 & 1.00 & 3.35 & 1.19 & 1.40 \\ 
\rowcolor{gray_tab}InsViE~\cite{insvie}      & 2B        & 720$\times$480  &   2.62 & 3.67 & 2.12 & 2.88 & 2.59 & 1.61 & 3.91 & 2.93 & 1.56 \\ 
\rowcolor{gray_tab}Lucy-Edit~\cite{lucyedit}   & 5B        & 1280$\times$704 & 3.42 & 3.81 & 3.17 & 4.43 & 2.85 & 2.92 & 2.98 & 3.75 & 3.58 \\
\rowcolor{gray_tab}ICVE~\cite{icve}        & 13B       & 384$\times$240  & 3.44 & 3.36 & 3.01 & 4.15 & 3.62 & 2.92 & 4.30 & 3.33 & 3.28 \\ 
\rowcolor{gray_tab}DITTO~\cite{ditto}       & 14B       & 832$\times$480  &  3.30 & 4.33 & 3.35 & 3.61 & 2.76 & 2.70 & 2.65 & 3.99 & 3.36 \\
\midrule
\rowcolor{blue_tab}OpenVE-Edit(Ours) & 5B        & 1280$\times$704  &    3.54 & 4.00 & 3.59 & 4.23 & 2.88 & 2.84 & 3.95 & 3.40 & 3.30 \\ 
        \toprule[0.1em]
        \end{tabular}
    }
    \label{tab:sup_internvl}
\end{table*}
\begin{table*}[htbp]
    \centering
    \renewcommand{\arraystretch}{1.0}
    \caption{Quantitative Comparison on \textbf{OpenVE-Bench} with Qwen3VL~\cite{qwen25vl}.}
    \label{tab:sup_qwen3vl}
    \resizebox{1\linewidth}{!}{
        \begin{tabular}{c|cc|c|cccccccc}
        \toprule[0.1em]
Methods     & \#Params. & \#Reso.  & Overall & Global Style & Background Change & Local Change & Local Remove & Local Add & Subtitle Edit & Creative Edit & Camera Edit \\
    \midrule
    \rowcolor{oran_tab}Runway Aleph      & -         & 1280$\times$720 & 4.49 & 4.41 & 4.40 & 4.31 & 4.64 & 4.36 & 4.21 & 4.81 & 4.70 \\ 
\rowcolor{gray_tab}VACE\cite{vace}         & 14B       & 1280$\times$720 & 3.01 & 3.46 & 2.81 & 2.47 & 3.99 & 1.76 & 4.41 & 2.17 & 3.09 \\
\rowcolor{gray_tab}OmniVideo~\cite{omnivideo}   & 1.3B      & 640$\times$352  & 1.68 & 1.37 & 2.02 & 1.65 & 1.42 & 1.93 & 1.79 & 1.23 & 1.80 \\
\rowcolor{gray_tab}InsViE~\cite{insvie}      & 2B        & 720$\times$480  &   3.25 & 3.63 & 2.68 & 2.82 & 3.56 & 2.25 & 4.77 & 3.36 & 3.61 \\
\rowcolor{gray_tab}Lucy-Edit~\cite{lucyedit}   & 5B        & 1280$\times$704 & 3.77 & 3.64 & 3.25 & 3.93 & 3.95 & 3.92 & 4.23 & 4.19 & 3.54 \\
\rowcolor{gray_tab}ICVE~\cite{icve}        & 13B       & 384$\times$240  & 3.76 & 3.87 & 3.51 & 3.87 & 4.50 & 3.77 & 4.68 & 3.54 & 2.84 \\
\rowcolor{gray_tab}DITTO~\cite{ditto}       & 14B       & 832$\times$480  & 3.44 & 4.48 & 3.52 & 2.89 & 3.53 & 2.48 & 3.69 & 4.14 & 3.33 \\
\midrule
\rowcolor{blue_tab}OpenVE-Edit(Ours) & 5B        & 1280$\times$704  &  3.89 & 4.24 & 4.10 & 3.80 & 3.50 & 3.41 & 3.98 & 3.71 & 3.25 \\
        \toprule[0.1em]
        \end{tabular}
    }
\end{table*}

\FloatBarrier

\prompt{Global Style}{
You are a data rater specializing in grading video style transfer edits. You will be given an input video, a reference style (image or video), and the styled result video. Your task is to evaluate the style transfer on a 5-point scale from three perspectives:

\textbf{Instruction Compliance}
1. Target style absent or clearly wrong.
2. Style shows in a few areas/frames only, or mixed with unrelated styles.
3. Key traits (palette, brushwork, texture) present but patchy or inconsistent across frames.
4. Style reproduced well across almost the whole video; only small local or brief temporal mismatches.
5. Full, faithful transfer: colour, texture, brushwork, and lighting all match the exemplar consistently over the entire duration and space of the video.

\textbf{Consistency \& Detail Fidelity}
1. Major objects, layout, or overall motion lost/distorted; original scene barely recognisable.
2. Main subject recognisable, but its size, perspective, motion, or key parts are clearly wrong/missing.
3. Overall structure and motion correct; some local warping, minor omissions, or slight motion jerkiness.
4. Nearly all geometry and motion intact; only slight, non-distracting deformation.
5. All objects, spatial relations, and motion are perfectly kept; only stylistic, harmless distortion.

\textbf{Visual Quality \& Stability}
1. Extreme flickering or "boiling" effects; the style is completely unstable frame-to-frame, making the video unwatchable.
2. Significant and distracting flickering or temporal inconsistency in style application.
3. Noticeable but tolerable flicker or texture "boiling", especially during motion.
4. Largely stable with only minor, subtle flickering visible in areas of complex motion or fine texture.
5. Perfectly stable and temporally coherent; the style appears "stuck" to the scene with no flickering.

\textbf{Note:} The scores for Consistency \& Detail Fidelity and Visual Quality \& Stability should not be higher than the Instruction Compliance score!!!

\textbf{Example Response Format}
Brief reasoning: A short explanation of the scores based on the criteria above, no more than 30 words.\\
Instruction Compliance: A number from 1 to 5.\\
Consistency \& Detail Fidelity: A number from 1 to 5.\\
Visual Quality \& Stability: A number from 1 to 5.\\
editing instruction is : \textless edit\_prompt\textgreater.

Below are the videos before and after editing:
}{prompt:gs}

\prompt{Background Change}{
You are a data rater specializing in grading video background editing. You will be given two videos (before and after editing) and the editing instruction. Your task is to evaluate the background change on a 5-point scale from three perspectives:

\textbf{Instruction Compliance}
1. No change, or background unrelated to prompt, or foreground also replaced/distorted.
2. Background partly replaced or wrong style/content; foreground noticeably altered.
3. Main background replaced but elements missing/extra, or faint spill onto subject edges.
4. Requested background fully present; foreground intact except minute artefacts or small prompt mismatch (e.g. colour tone).
5. Background exactly matches prompt (content, style, placement); all foreground pixels untouched.

\textbf{Consistency \& Detail Fidelity}
1. Large tearing, posterisation, or significant temporal artifacts like flickering, jittering edges; edit area obvious at a glance.
2. Clear cut-out halos, colour-resolution gap, or obvious edge 'boiling' (instability) over time.
3. Blend acceptable but visible on closer look: slight edge blur, or minor temporal instability (e.g., shimmer).
4. Nearly invisible seams; edges are stable across motion, textures aligned, only minor issues when zoomed in.
5. Indistinguishable composite: edges, textures, resolution and colour grading are perfectly continuous and stable throughout the video's duration.

\textbf{Visual Quality \& Stability}
1. Severe mismatch: wrong horizon, conflicting light, floating subject, or background remains static during camera movement (no parallax).
2. Noticeable inconsistencies in light or scale; incorrect perspective shifts during motion.
3. Overall believable; small errors in shadow, perspective, or minor motion tracking flaws.
4. Lighting, scale, and depth well matched; background perspective and scale track convincingly with camera motion.
5. Physically flawless: foreground and new background share coherent light, shadows, perspective, and atmospheric depth throughout all subject and camera motion, enhancing overall realism.

The second and third score should no higher than first score!!!

\textbf{Example Response Format:}
Brief reasoning: A short explanation of the score based on the criteria above, no more than 20 words.\\
Instruction Compliance: A number from 1 to 5.\\
Consistency \& Detail Fidelity: A number from 1 to 5.\\
Visual Quality \& Stability: A number from 1 to 5.\\
editing instruction is : \textless edit\_prompt\textgreater.

Below are the videos before and after editing:
}{prompt:bc}

\prompt{Local Change}{
You are a data rater specializing in grading video replacement edits. You will be given two videos (before and after editing) and the corresponding editing instructions. Your task is to evaluate the replacement editing effect on a 5-point scale from three perspectives, paying close attention to temporal consistency (how the edit holds up over time and with motion).

\textbf{Instruction Compliance}
1. Target not replaced, or an unrelated object/part of the video edited.
2. Only part of the target replaced (e.g., in only a few frames), or wrong class/description used.
3. Target largely replaced but other objects altered, remnants visible across frames, or count/position clearly wrong.
4. Correct object fully replaced for the entire duration; only minor attribute errors (colour, size, etc.).
5. Perfect replacement: all and only the specified objects replaced for the entire duration; new objects’ class, number, position, scale, pose, motion and detail exactly match the prompt.

\textbf{Consistency \& Detail Fidelity}
1. Video heavily broken or new object deformed / flickers uncontrollably / jitters erratically.
2. Obvious seams/edges that flicker or move unnaturally; strong mismatch in resolution or colour that is inconsistent across frames; background not restored or is unstable.
3. Basic style similar, but lighting or palette clashes are inconsistent as the video plays; fuzzy edges, noise or minor flickering/jittering are noticeable.
4. Style almost uniform and stable; tiny temporal artefacts (e.g., edge shimmer) visible only on close, frame-by-frame inspection; casual viewers see no edit.
5. Completely seamless and temporally stable; new objects blend fully with the scene in every frame, edit area undetectable.

\textbf{Visual Quality \& Stability}
1. Floating or sliding unnaturally (poor motion tracking), severe perspective/light errors inconsistent with camera/object movement; background heavily warped or unstable.
2. Missing or static shadows/reflections that do not move with the object/light; poor occlusion; new object’s motion clearly mismatches scene motion.
3. Lighting, perspective and interactions mostly correct but with minor inconsistencies over time; motion tracking has small, tolerable drifts.
4. New object's motion is well-tracked and it interacts realistically with the scene (shadows, reflections) and preserves existing details throughout the video.
5. Physically and dynamically flawless: motion, perspective, shadows, and reflections are perfectly integrated and move correctly with the scene and camera in every frame; background untouched and stable.

The second and third score should no higher than first score!!!

\textbf{Example Response Format:}
Brief reasoning: A short explanation of the score based on the criteria above, no more than 20 words.\\
Instruction Compliance: A number from 1 to 5.\\
Consistency \& Detail Fidelity: A number from 1 to 5.\\
Visual Quality \& Stability: A number from 1 to 5.\\
editing instruction is : \textless edit\_prompt\textgreater.

Below are the videos before and after editing:
}{prompt:lc}

\prompt{Local Remove}{
You are a data rater specializing in grading video object removal editing. You will be given two videos (before and after editing) and the corresponding editing instructions. Your task is to evaluate the edit quality on a 5-point scale from three perspectives:

\textbf{Instruction Compliance}
1. No edit performed, the video is corrupted, or the edit is completely wrong.
2. Wrong object/class removed, or target only partially removed, or an unrelated object is also removed.
3. Correct object removed, but with significant errors: unintended objects are also removed, OR significant fragments/ghosting of the target remain.
4. The correct object is removed; only minor issues like a few tiny fragments remaining or tiny, unintended background items being affected.
5. Perfect: All and only the requested objects are removed as instructed; every other element is untouched.

\textbf{Visual Quality \& Stability}
1. Video is badly broken, full of artefacts, or shows severe flickering/jittering throughout.
2. Obvious erase marks or "smudges"; the inpainted background's style, resolution, or palette strongly mismatches; the edited region jitters or appears static against a moving background.
3. General style is similar, but the inpainted background's lighting/colours clearly clash or are inconsistent across frames; noticeable temporal disharmony.
4. Style is almost uniform; minor edge issues around the removed area or slight temporal instability (e.g., minor flicker) visible only on close inspection.
5. Perfectly seamless; the removal is temporally stable and visually indistinguishable from a clean background.

\textbf{Consistency \& Detail Fidelity}
1. Key original elements are blocked by poor inpainting; the background is heavily distorted or hallucinates incorrect structures; motion is completely wrong (e.g., a static patch in a moving scene).
2. The inpainted background visibly shifts, jitters, or is poorly reconstructed over time, failing to match the original scene's motion.
3. Background reconstruction is mostly correct and consistent; remaining flaws are small and acceptable; background changes are localized and stable.
4. No loss of original detail around the removed area; background reconstruction is clean, stable, and respects the scene's geometry and motion.
5. The background is essentially untouched and stable; the inpainted area perfectly matches the surrounding content's motion, texture, and detail over time.

The second and third score should no higher than first score!!!

\textbf{Example Response Format:}
Brief reasoning: A short explanation of the score based on the criteria above, no more than 20 words.\\
Instruction Compliance: A number from 1 to 5.\\
Visual Quality \& Stability: A number from 1 to 5.\\
Consistency \& Detail Fidelity: A number from 1 to 5.\\
editing instruction is : \textless edit\_prompt\textgreater.

Below are the videos before and after editing:
}{prompt:lr}

\prompt{Local Add}{
You are a data rater specializing in grading video object addition editing. You will be given two videos (before and after editing) and the corresponding editing instructions. Your task is to evaluate the edit quality on a 5-point scale from three perspectives:

\textbf{Instruction Compliance}
1. No edit performed, the video is corrupted, or the edit is completely wrong.
2. Wrong object/class added, or target only partially added, or an unrelated object is also added.
3. Correct object added, but with significant errors: key attributes (e.g., position, colour, count, size) are wrong.
4. The correct object is added with main attributes correct; only minor details are off (e.g., slight colour mismatch, minor position error).
5. Perfect: All and only the requested objects are added as instructed; every other element is untouched.

\textbf{Visual Quality \& Stability}
1. Video is badly broken, full of artefacts, or shows severe flickering/jittering throughout.
2. Obvious paste marks; style, resolution, or palette of the added object strongly mismatches; the added region jitters or appears static against a moving background.
3. General style is similar, but lighting/colours on the added object clearly clash or are inconsistent across frames; noticeable temporal disharmony.
4. Style is almost uniform; minor edge issues around the added object or slight temporal instability (e.g., minor flicker) visible only on close inspection.
5. Perfectly seamless; the edit is temporally stable and visually indistinguishable from the original video's content and motion.

\textbf{Consistency \& Detail Fidelity}
1. Severe physical errors (e.g., the added object floats, has wrong perspective/lighting); key original elements are blocked; motion of the added object is completely wrong.
2. Contact with surfaces, occlusion by other objects, or motion of the added object is handled poorly.
3. Lighting, perspective, and motion of the added object are mostly correct and consistent with the scene; remaining flaws are small and acceptable.
4. Shadows, reflections, and material response from the added object are believable and move correctly with the scene; no loss of original detail.
5. Edit enhances overall realism: the added object has precise highlights, shadows, and motion effects that are temporally coherent and perfectly integrated.

The second and third score should no higher than first score!!!

\textbf{Example Response Format:}
Brief reasoning: A short explanation of the score based on the criteria above, no more than 20 words.\\
Instruction Compliance: A number from 1 to 5.\\
Visual Quality \& Stability: A number from 1 to 5.\\
Consistency \& Detail Fidelity: A number from 1 to 5.\\
editing instruction is : \textless edit\_prompt\textgreater.

Below are the videos before and after editing:
}{prompt:la}

\prompt{Subtitles Edit}{
You are a data rater specializing in grading instruction-following subtitle edits. You will be given two videos (before and after editing) and the corresponding editing instructions. Your task is to evaluate the subtitle edit on a 5-point scale from three perspectives:

\textbf{Instruction Compliance}
1. Target subtitle not added/removed/replaced, or wrong subtitle affected.
2. Right action (add/remove/replace) but with incorrect content; only part of the edit is done; other subtitles are also altered.
3. Mainly correct action and content, yet with significant spelling/grammar errors, or minor unintended edits to other subtitles.
4. Correct action performed on the right subtitle; content is correct with only minor inaccuracies (e.g., small typos, punctuation errors).
5. Exactly and only the requested subtitle(s) are added/removed/replaced; content matches the prompt perfectly; zero unintended edits.

\textbf{Visual Quality \& Stability}
1. Completely fails to follow specified attributes (e.g., wrong position, wrong color). If attributes are not specified, the chosen ones make the subtitle unreadable or are extremely disruptive.
2. Major deviation from specified attributes (e.g., requested bottom, placed on top). If not specified, chosen attributes are clearly wrong and distracting (e.g., obscures key visuals).
3. Follows the general direction of specified attributes but with significant errors (e.g., correct side but wrong exact position). If not specified, chosen attributes are acceptable but noticeably inconsistent.
4. Follows specified attributes with only minor inaccuracies (e.g., slightly off-center, minor deviation in font/color). If not specified, chosen attributes are highly appropriate with only minor flaws.
5. All specified attributes (position, font, color, etc.) are matched perfectly. If attributes are not specified, the chosen ones are perfectly consistent with existing subtitles or professional standards.

\textbf{Consistency \& Detail Fidelity}
1. Massive collateral damage: background video is heavily corrupted/glitched, or other non-target subtitles are wrongly deleted/altered.
2. Noticeable collateral damage: visible artifacts, distortion, or color shifts in the background video; other subtitles are visibly affected.
3. Minor unintended effects: slight and localized visual artifacts in the background, or minor, non-critical changes to adjacent subtitles' appearance/timing.
4. Almost perfect preservation: only extremely subtle artifacts in the video frame, visible only upon close inspection; all other subtitles are untouched.
5. Perfect preservation: the edit is perfectly isolated; the background video and all other subtitles remain 100

The second and third score should no higher than first score!!!

\textbf{Example Response Format:}
Brief reasoning: A short explanation of the score based on the criteria above, no more than 2, no more than 20 words.\\
Instruction Compliance: A number from 1 to 5.\\
Visual Quality \& Stability: A number from 1 to 5.\\
Consistency \& Detail Fidelity: A number from 1 to 5.\\
editing instruction is : \textless edit\_prompt\textgreater.

Below are the videos before and after editing:
}{prompt:se}

\prompt{Camera Multi-shot Edit}{
You are a data rater specializing in grading camera shot type alteration edits. You will be given two videos (before and after editing) and the corresponding editing instructions. Your task is to evaluate the camera shot change on a 5-point scale from three perspectives:

\textbf{Instruction Compliance}
1. The shot type is not changed, or changed to a completely wrong type (e.g., requested close-up, but got a long shot).
2. The direction of the shot change is correct (e.g., zoomed in for a close-up), but the degree is wrong (e.g., a medium shot instead of a close-up).
3. The shot type is generally correct, but the framing is poor, cutting off important parts of the subject or being poorly centered.
4. The shot type and framing are correct, with only minor inaccuracies in composition.
5. The video is perfectly transformed into the requested shot type (long, medium, or close-up) with ideal framing of the subject.

\textbf{Visual Quality \& Stability}
1. Massive distortion, glitches, warping, or heavy noise; the edited video is unusable.
2. Significant and distracting jitter, shimmering, or warping is visible throughout the video, making the shot feel unstable.
3. Minor but noticeable visual flaws, such as slight edge distortion or a subtle "breathing" effect in the frame.
4. The video is stable and clear, with only very slight, almost unnoticeable artifacts upon close inspection.
5. The resulting shot is perfectly stable and clear, with no digital artifacts, distortion, or jitter. It looks as if it were originally filmed with that shot type.

\textbf{Consistency \& Detail Fidelity}
1. The subject, background, or action in the edited video is completely different from the original video; a total failure of consistency.
2. The main subject is the same, but their action, the background, or the lighting is drastically and illogically changed compared to the original video.
3. The scene is generally consistent, but there are noticeable continuity errors (e.g., an object disappears, the subject's pose changes unnaturally).
4. The subject, action, and environment are highly consistent with the original video. Original details are well-preserved with only minor, hard-to-spot discrepancies.
5. Perfect consistency; the edited video perfectly preserves the subject, lighting, background, and continuity of action from the original video, creating the illusion of the same scene captured from a different camera position.

The second and third score should no higher than first score!!!

\textbf{Example Response Format:}
Brief reasoning: A short explanation of the score based on the criteria above, no more than 20 words.\\
Instruction Compliance: A number from 1 to 5.\\
Visual Quality \& Stability: A number from 1 to 5.\\
Consistency \& Detail Fidelity: A number from 1 to 5.\\
editing instruction is : \textless edit\_prompt\textgreater.

Below are the videos before and after editing:
}{prompt:cme}

\prompt{Creative Edit}{
You are a data rater specializing in grading instruction-following creative video edits. You will be given two videos (before and after editing) and the corresponding editing instructions. Your task is to evaluate the creative edit on a 5-point scale from three perspectives:

\textbf{Instruction Compliance}
1. The instruction is completely ignored or the edit is irrelevant to the prompt.
2. The edit attempts the instruction but fundamentally fails; the core subject, style, or action is wrong or only applied for a brief moment.
3. The edit generally follows the instruction, but with major deviations in style, motion, or concept; the effect is highly inconsistent over time.
4. The edit successfully executes the instruction with only minor inaccuracies in style, motion, or detail; the result is temporally consistent.
5. The edit perfectly and creatively interprets and executes the instruction throughout the video's duration, fully achieving the intended creative goal.

\textbf{Visual Quality \& Stability}
1. Massive flickering, strobing, or artifacts that make the video unwatchable; edited elements are completely disjointed from the scene.
2. Obvious temporal inconsistency (e.g., style flickers on/off), clear visual boundaries or seams; mismatched color/lighting between frames.
3. The edit is mostly stable, but with noticeable "boiling" or "shimmering" in textures/styles; minor jitter or softness on edges.
4. The edit is very stable and well-integrated; only slight, hard-to-spot artifacts or flickering are present, motion is smooth.
5. Perfectly stable and seamless integration; the edit feels like part of the original footage with no detectable flickering, jitter, or discontinuities.

\textbf{Consistency \& Detail Fidelity}
1. Complete break from physical laws; added objects have no correct lighting/shadows, move unnaturally; original video details are heavily degraded.
2. Major physical inconsistencies; shadows/reflections are static or move incorrectly; motion of edits doesn't match camera movement; original background is warped.
3. Physics and lighting are generally believable but with minor flaws (e.g., a shadow is slightly off); unedited parts of the video are mostly preserved.
4. Edited elements interact realistically with the scene's lighting, motion, and perspective; original video details are well-preserved.
5. High degree of physical realism and integration; motion, lighting, and physics of the edits are indistinguishable from a real recording; original details are perfectly maintained.

The second and third score should no higher than first score!!!

\textbf{Example Response Format:}
Brief reasoning: A short explanation of the score based on the criteria above, no more than 20 words.\\
Instruction Compliance: A number from 1 to 5.\\
Visual Quality \& Stability: A number from 1 to 5.\\
Consistency \& Detail Fidelity: A number from 1 to 5.\\
editing instruction is : \textless edit\_prompt\textgreater.

Below are the videos before and after editing:
}{prompt:cre}